\documentclass[journal]{IEEEtran}

\ifCLASSINFOpdf
  \usepackage[pdftex]{graphicx}
  \DeclareGraphicsExtensions{.pdf,.jpeg,.png}
\else
  \usepackage[dvips]{graphicx}
  \DeclareGraphicsExtensions{.pdf,.jpeg,.png}
\fi

\usepackage{enumitem}
\usepackage{amssymb}
\usepackage{todonotes}
\usepackage{pifont}

\usepackage{titling}  
\usepackage{placeins}  

\usepackage{url}
\usepackage{color, colortbl}
\definecolor{lightgray}{rgb}{0.92,0.92,0.92}
\newcolumntype{g}{>{\columncolor{lightgray}}c}
\usepackage{booktabs}
\usepackage{multirow}  
\usepackage{rotating}  
\newcommand{\hcell}[1]{\cellcolor[HTML]{DAE8FC}{#1}}  
\newcolumntype{?}{!{\vrule width 1.5pt}}  
\usepackage{amsmath}
\usepackage{dsfont}

\newcommand{\mathimage}[1]{\mathbf{#1}}

\newcommand{\mathscal}[1]{\lowercase{\textit{#1}}}
\newcommand{\mathtensor}[1]{\mathrm{\uppercase{#1}}}
\newcommand{\mathdistrib}[1]{\mathcal{#1}}
\usepackage{upgreek}  
\newcommand{\norm}[1]{\left\lVert#1\right\rVert} 
\newcommand{\purpletext}[1]{\color{purple}{#1}\color{black}}

\usepackage{algorithmic}

\usepackage{pifont}
\newcommand{\cmark}{\ding{51}}
\newcommand{\xmark}{\ding{55}}

\begin{document}


\title{Learning to Segment from Scribbles using Multi-scale Adversarial Attention Gates}

\author{Gabriele~Valvano, Andrea~Leo, Sotirios~A.~Tsaftaris
    \thanks{G. Valvano (email: gabriele.valvano@imtlucca.it) and A. Leo are with IMT School for Advanced Studies Lucca, Lucca 55100 LU, Italy. G. Valvano is also with School of Engineering, University of Edinburgh, Edinburgh EH9 3FB, UK. S. A. Tsaftaris is with School of Engineering, University of Edinburgh, Edinburgh EH9 3FB, UK.}
    \thanks{This work was supported by the Erasmus+ programme of the European Union. S.A. Tsaftaris acknowledges the support of the Royal Academy of Engineering and the Research Chairs and Senior Research Fellowships scheme. We thank NVIDIA for donating the GPU used for this research.}
}
\date{}

\maketitle

\begin{abstract}

Large, fine-grained image segmentation datasets, annotated at pixel-level, are difficult to obtain, particularly in medical imaging, where annotations also require expert knowledge.
Weakly-supervised learning can train models by relying on weaker forms of annotation, such as scribbles. 
Here, we learn to segment using scribble annotations in an adversarial game. With unpaired segmentation masks, we train a multi-scale GAN to generate realistic segmentation masks at multiple resolutions, while we use scribbles to learn their correct position in the image. 
Central to the model's success is a novel attention gating mechanism, which we condition with adversarial signals to act as a shape prior, resulting in better object localization at multiple scales. Subject to adversarial conditioning, the segmentor learns attention maps that are semantic, suppress the noisy activations outside the objects, and reduce the vanishing gradient problem in the deeper layers of the segmentor.
We evaluated our model on several medical (ACDC, LVSC, CHAOS) and non-medical (PPSS) datasets, and we report performance levels matching those achieved by models trained with fully annotated segmentation masks. 
We also demonstrate extensions in a variety of settings: semi-supervised learning; combining multiple scribble sources (a crowdsourcing scenario) and multi-task learning (combining scribble and mask supervision).
We release expert-made scribble annotations for the ACDC dataset, and the code used for the experiments, at \purpletext{\url{https://vios-s.github.io/multiscale-adversarial-attention-gates}}.
\footnote{\purpletext{Manuscript accepted at: IEEE Transaction on Medical Imaging, 2021.}}
\end{abstract}


\IEEEpeerreviewmaketitle

\section{Introduction}
Convolutional Neural Networks (CNNs) have obtained impressive results in computer vision. However, their ability to generalize on new examples is strongly dependent on the amount of training data, thus limiting their applicability when annotations are scarce. 
There has been a considerable effort to exploit semi-supervised and weakly-supervised strategies.
For semantic segmentation, semi-supervised learning (SSL) aims to use unlabeled images, generally easier to collect, together with some fully annotated image-segmentation pairs~\cite{chapelle2009semi, cheplygina2019not}.
However, the information inside unlabeled data can improve CNNs only under specific assumptions~\cite{chapelle2009semi}, and SSL requires representative image-segmentation pairs being available. 

Alternatively, weakly-supervised approaches~\cite{zhou2019prior, khoreva2017simple, can2018learning, souly2017semi} attempt to train models relying only on weak annotations (e.g., image-level labels, sparse pixel annotations, or noisy annotations~\cite{tajbakhsh2020embracing}), that should be considerably easier to obtain. Thus, building large-scale annotated datasets becomes feasible and the generalization capability of the model per annotation effort can dramatically increase: e.g., 15 times more bounding boxes can be annotated within the same time compared to segmentation masks~\cite{lin2014microsoft}. Among weak annotations, scribbles are of particular interest for medical image segmentation, because they are easier to generate and well suited for annotating nested structures~\cite{can2018learning}. 
Unfortunately, learning from weak annotations does not provide a supervisory signal as strong as one obtained from fine-grained per-pixel segmentation masks, and training CNNs is harder. Thus, improved training strategies can enable remarkable gains with weaker forms of annotations.

\begin{figure}[t]
    \centering
    \includegraphics[width=0.48\textwidth]{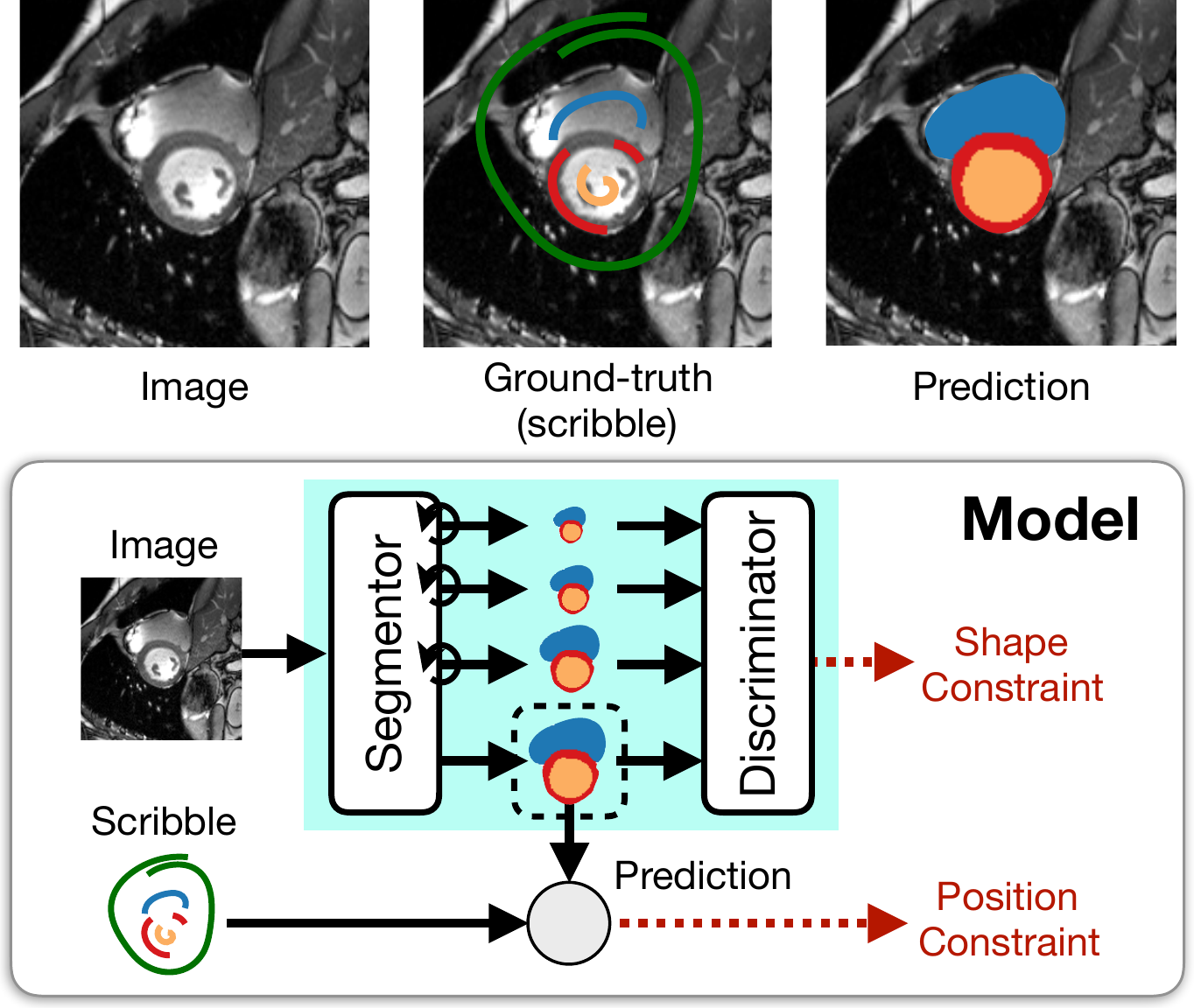}
    \caption{%
    In an adversarial game, our model learns to generate segmentation masks that look realistic at multiple scales and overlap with the available scribble annotations. Loopy arrows in the figure, on the segmentor, represent the proposed attention gates, which under adversarial conditioning suppress irrelevant information in the extracted features maps.%
    }
    \label{fig:method}
\end{figure}
\subsection{Overview of the proposed approach}
In this paper, we introduce a novel training strategy in the context of weakly supervised learning for multi-part segmentation. We train a model for semantic segmentation using scribbles, shaping the training procedure as an adversarial game~\cite{goodfellow2014generative} between a conditional mask generator (the segmentor) and a discriminator. We obtain segmentation performance comparable to when training the segmentor with full segmentation masks. We demonstrate this for the segmentation of the heart, abdominal organs, and human pose parts.
Our uniqueness is that we use adversarial feedback at all scales, coupling the generator with a multi-scale discriminator. But, differently from other multi-scale GANs~\cite{denton2015deep,karras2017progressive,luo2018macro}, our generator includes customized attention gates, i.e. modules that automatically produce soft region proposals in the feature maps, highlighting the salient information inside of them.
Differently from the attention gates presented in~\cite{schlemper2019attention} ours are conditioned by the adversarial signals, which enforce a stronger object localization in the image.
Moreover, differently from other multi-scale GANs~\cite{denton2015deep,karras2017progressive,luo2018macro} we use a single discriminator rather than multiple ones, thus reducing the computational cost whilst retaining their advantages in semantic segmentation.

The discriminator, acting as a learned shape prior, is trained on a set of segmentation masks, obtained from a different data source\footnote{We simulate a realistic clinical setting, where the unpaired masks can be obtained from a different modality or acquisition protocol \cite{larrazabal2020post,painchaud2020cardiac}.} and is thus unpaired. We drive the segmentor to generate accurate segmentations from the input images, while satisfying the multi-scale shape prior learned by the discriminator. 
We encourage a tight multi-level interaction between segmentor and discriminator introducing \emph{Adversarial Attention Gating}, an effective attention strategy that, subject to adversarial conditioning, i) encourages the segmentor to predict masks satisfying multi-resolution shape priors; and ii) forces the segmentor to train deeper layers better.
Finally, we also penalize the segmentor when it predicts segmentations that do not overlap with the available scribbles, pushing it to learn the correct mapping from images to label maps. 

We summarize the contributions of the paper %
as follows:
\begin{itemize}
    \item We use scribble annotations to learn semantic segmentation during a multi-scale adversarial game.
    \item We introduce Adversarial Attention Gates (AAGs): effective prior-driven attention gates that force the segmentor to localize objects in the image. Subject to adversarial gradients, AAGs also encourage a better training of deeper layers in the segmentor.
    \item We obtain state-of-the-art performance compared to other scribble-supervised models on several popular medical datasets (ACDC~\cite{bernard2018deep}, LVSC~\cite{suinesiaputra2014collaborative} and CHAOS~\cite{chaos}) and computer vision data (PPSS~\cite{luo2013pedestrian}).
    \item We investigate diverse learning scenarios, such as: learning from different extents of weak annotations (i.e., semi-supervised learning); learning from multiple scribbles per image (and thus simulating a crowdsourcing setting); and finally learning also with few strong supervision pairs of segmentation masks and images (i.e., multi-task learning).
    \item Lastly, we compare our model, trained on scribbles, with a few-shot learning method trained with densely annotated segmentation masks, and show the advantage of collecting large-scale weakly annotated datasets.
\end{itemize}

\section{Related Work}
A large body of research aimed at developing learning algorithms that rely less on high-quality annotations~\cite{cheplygina2019not,tajbakhsh2020embracing}. Below, we briefly review recent weakly supervised methods that use scribbles to learn image segmentation. Then, we discuss what are the advantages of our adversarial setup compared to other multi-scale GANs. Finally, we discuss the difference between the attention gates that are an integral part of our segmentor and other canonical attention modules.

\subsection{Learning from Scribbles}
Scribbles are sparse annotations that have been successfully used for semantic segmentation, reporting near full-supervision accuracy in computer vision and medical image analysis. However, scribbles lack information on the object structure, and they are limited by the uncertainty of unlabelled pixels, which makes training CNNs harder, especially in boundary regions~\cite{lin2016scribblesup}. %
For this reason, many approaches have tried to expand scribble annotations by assigning the same class to pixels with similar intensity and nearby position~\cite{lin2016scribblesup, ji2019scribble}. At first, these approaches relabel the training set propagating annotations from the scribbles to the adjacent pixels using graph-based methods. Then, they train a CNN on the new label maps. 
A recent variant has been introduced by Can \textit{et al.}~\cite{can2018learning}, who suggest estimating the class of unlabelled pixels via a learned two-step procedure. At first, they train a CNN directly with scribbles; then, they relabel the training set by refining the CNN predictions with Conditional Random Fields (CRF); finally, they retrain the CNN on the new annotations.

The major limitation of the aforementioned approaches is relying on dataset relabeling, which can be time-consuming and is prone to errors that can be propagated to the models during training. Thus, many authors~\cite{can2018learning,tang2018regularized} have investigated alternatives that avoid this step, post-processing the model predictions with CRF~\cite{chen2017deeplab} or introducing CRF as a trainable layer~\cite{zheng2015conditional}. Tang \textit{et al.}~\cite{tang2018regularized} have also demonstrated the possibility to substitute the CRF-based refining step, directly training a segmentor with a CRF-based loss regulariser. 

Similarly, here we propose a method that avoids the data relabeling step. We train our model to directly learn a mapping from images to segmentation masks, and we remove expensive CRF-based post-processing. We cope with unlabelled regions of the image introducing a multi-scale adversarial loss which, differently from the loss introduced by Tang \textit{et al.}~\cite{tang2018regularized}, does not rely on CRF, and can handle both long-range and short-range inconsistencies in the predicted masks.

Concurrent to our work, Zhang \textit{et al.}~\cite{zhang2020accl} recently introduced a method that learns to segment images from scribbles using an adversarial shape prior. However, they suggest using a PatchGAN~\cite{isola2017image} discriminator, which only focuses on \textit{local} properties of the generated segmentations, while we introduce a method that focuses on both \textit{local} and \textit{global} aspects.

\subsection{Shape Priors in Deep Learning for Medical Imaging}
In semantic segmentation, there has been considerable interest in incorporating prior knowledge about organ shapes to obtain more accurate and plausible results~\cite{nosrati2016incorporating}. Below, we summarise recent work on shape priors in Deep Learning.

Recently, Clough \textit{et al.} used Persistent Homology to enforce shape priors in medical image segmentation~\cite{clough2019topological}. Oktay \textit{et al.}~\cite{oktay2017anatomically} demonstrated that we can learn a data-driven shape prior with a convolutional autoencoder trained on unpaired segmentation masks, and it can be used as regulariser to train a segmentor. Dalca \textit{et al.}~\cite{dalca2018anatomical} suggested learning the shape prior with a variational autoencoder (VAE)~\cite{kingma2014auto}, and then share part of the VAE weights with a segmentor. Other approaches included shape priors as post-processing, regularising the training~\cite{yue2019cardiac}, or adjusting predictions at inference, using VAEs~\cite{painchaud2019cardiac} or denoising autoencoders~\cite{larrazabal2020post}. 
Kervadec \textit{et al.}~\cite{kervadec2019constrained} suggested introducing size information as a differentiable penalty, during training.
Alternatively, Dalca \textit{et al.}~\cite{dalca2019unsupervised} proposed to learn to warp a segmentation atlas.
Other methods~\cite{kohl2018probabilistic, baumgartner2019phiseg} proved that image segmentation has intrinsic uncertainty, which can be reflected in the learned shape prior. Finally, a body of literature showed that decoupling (disentangling) object shapes and appearance is beneficial in a lack of data~\cite{chartsias2019disentangled, yang2019unsupervised}, as well as using temporal consistency constraints on the object shapes dynamics~\cite{valvano2019temporal}.

Herein, we will focus on a particular type of shape prior, learned by a multi-scale GAN from unpaired segmentation masks. Particularly, we use an adversarial loss during training and avoid expensive post-processing of the predicted masks.

\subsection{Multi-scale GANs}
Herein, we use the generator as a segmentor, which we train to predict realistic segmentation masks at multiple scales. Recently, other methods introduced multi-scale adversarial losses for segmentation. For example, Xue \textit{et al.}~\cite{xue2018segan} proposed to use the discriminator as a critic, measuring the $\ell_1$-distance between \textit{real} and \textit{fake} inputs in features space, at multiple resolution levels. In particular, pairs of real and fake inputs consist in the Hadamard product between an image and the associated ground truth or predicted segmentation mask, respectively. Also Luo \textit{et al.}~\cite{luo2018macro} separated \textit{real} from \textit{fake} input pairs at multiple scales, using two separate discriminators (one working at high, one at low resolution) to distinguish the image concatenation with the associated ground truth or predicted segmentation, respectively.

Unfortunately, these approaches rely on image-segmentation pairs to train the discriminator. Thus, training the segmentor with unlabelled, or weakly annotated data is not possible. Instead, we train a discriminator using \emph{only} masks, making the model suitable for semi- and weakly-supervised learning.
Also, contrarily to~\cite{luo2018macro}, we use a single multi-scale discriminator rather than two, keeping the computational cost lower.

Finally, while previous approaches use multi-scale GANs with strong annotations, this is, to the best of our knowledge, the first work to explore their use in weakly-supervised learning. Furthermore, we alter the canonical interplay between discriminator and segmentor to improve the object localization in the image, that we obtain with a novel adversarial conditioning of the attention maps learned by the segmentor.

\subsection{Attention Gates}
Due to the ability to suppress irrelevant and ambiguous information, attention gates have become an integral part of many sequence modeling~\cite{vaswani2017attention} and image classification~\cite{Jetley2018} frameworks. 
Recently, they have also been successfully employed for segmentation~\cite{schlemper2019attention,oktay2018attention,wang2018deep,sinha2020multi, fu2019dual}, along with the claim that gating helps to detect desired objects. 
However, standard approaches don't incorporate any explicit constraint in the learned attention maps, which are generally predicted by the neural network autonomously. 
On the contrary, we show that conditioning the attention maps to be semantic, i.e. able to localize and distinguish separate objects, considerably boosts the segmentation performance. Herein, we introduce a novel attention module named Adversarial Attention Gate (AAG), whose learning is conditioned by a discriminator.

\section{Proposed Approach}
In this section, we first describe the adopted notation, and then we present a general overview of the proposed method. Finally, we detail model architectures and training objectives.

\subsubsection*{Notation} For the remainder, we will use italic lowercase letters to denote scalars $\mathscal{s}$. Two-dimensional images (matrices) will be denoted with bold lowercase letters, as $\mathimage{x} \in \mathds{R}^{n \times m}$, where $n, m \in \mathbb{N}$ are scalars denoting dimensions. Tensors $\mathtensor{T} \in \mathds{R}^{r \times s \times t}$ are denoted as uppercase letters, where $r, s, t \in \mathbb{N}$. Finally, capital Greek letters will denote functions $\upphi(\cdot)$. 

We will assume a weakly supervised setting, where we have access to: i) image-scribble pairs $(\mathimage{x}, \mathimage{y_{s}})$, being $\mathimage{x}$ the image and $\mathimage{y_{s}}$ the associated scribble; ii) unlabelled images; and iii) a set of segmentation masks $\mathimage{y}$ unrelated to any of the images.\footnote{In Section~\ref{subsec:multitask}, we will also investigate a mixed setting, where we additionally have: iv) pairs of image-segmentation masks $(\mathimage{x}, \mathimage{y})$.}

\begin{figure}[t]
    \centering
    \includegraphics[width=0.48\textwidth]{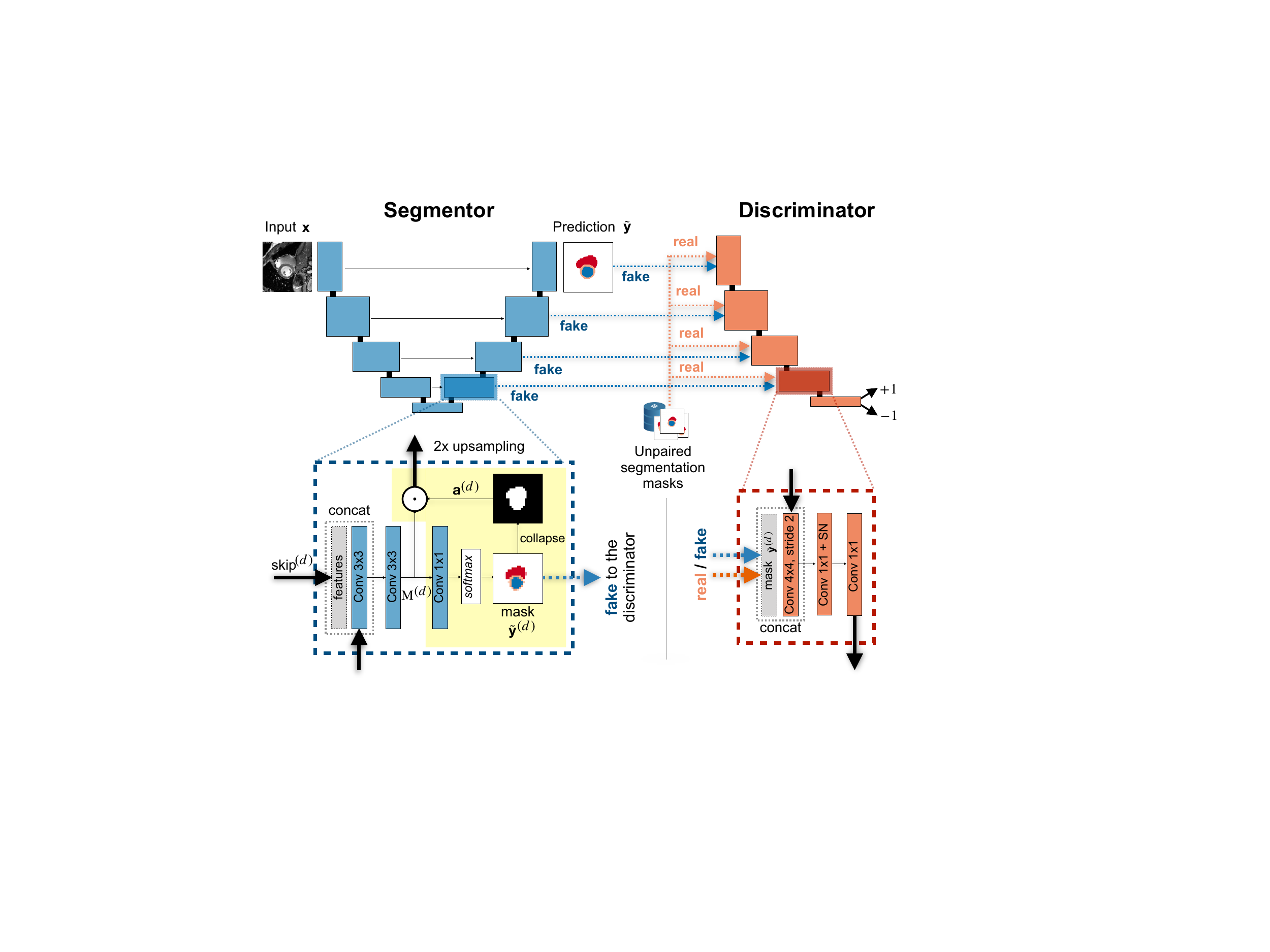}
    \caption{Model architectures. Top: segmentor and discriminator interact at multiple scales. Bottom: convolutional blocks detail. In yellow background, the Adversarial Attention Gate (AAG).}
    \label{fig:architectures}
\end{figure}
\begin{figure}
    \centering
    \includegraphics[width=0.48\textwidth]{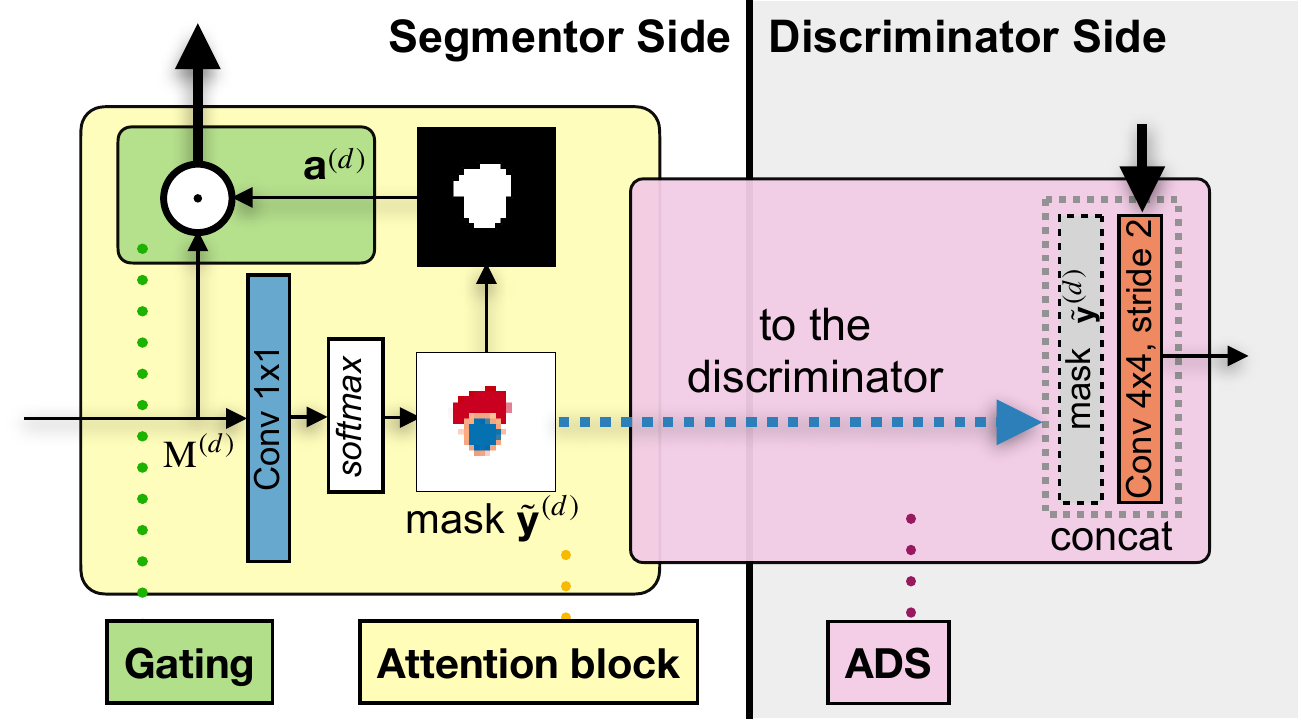}
    \caption{
    Adversarial Attention Gates consist of an attention block (yellow background) pairing Adversarial Deep Supervision (ADS, obtained via the connection in pink background) and a multiplicative gating operation (green background).}
    \label{fig:aag_components}
\end{figure}

\subsection{Method Overview} 
We formulate the training of a CNN with weak supervision (i.e., scribbles) as an adversarial game. Particularly, we use an adversarial discriminator to learn a multi-resolution shape prior, and we enforce a mask generator, or segmentor, to satisfy it, supported by the purposely designed adversarial attention gates. Critically, AAGs localize the objects to segment at multiple resolution levels and suppress noisy activations in the remaining parts of the image (see Fig.~\ref{fig:architectures}).

In detail, we jointly train a multi-scale segmentor $\Sigma(\cdot)$ and a multi-scale adversarial discriminator $\Delta(\cdot)$.
$\Sigma(\cdot)$ is supervisedly trained to predict segmentation masks $\tilde{\mathimage{y}} = \Sigma(\mathimage{x})$ that overlap with the scribble annotations, when available. Meanwhile, $\Delta(\cdot)$ learns to distinguish real segmentation masks from those (fake) predicted by the segmentor (i.e., $\Delta(\mathimage{y})$ vs $\Delta(\tilde{\mathimage{y}})$)~\cite{goodfellow2014generative}, at multiple scales. 
We model both $\Sigma(\cdot)$ and $\Delta(\cdot)$ as CNNs. 

In principle, other models can be used to learn multi-scale shape priors, as multi-scale VAEs~\cite{baumgartner2019phiseg, vahdat2020nvae}. We use GANs because they can be trained together with the segmentor in an adversarial game. The potential of using multi-scale VAEs in weakly supervised segmentation learning is an open research problem, which we leave for future work.

\subsection{Architectures}\label{subsec:architectures}
\paragraph*{Segmentor $\Sigma(\cdot)$} We modify a UNet~\cite{ronneberger2015u} to include AAG modules in the decoder and to allow collaborative training between segmentor and discriminator at multiple scales (Fig.~\ref{fig:architectures}).
We leave the UNet encoder as in the original framework, allowing to extract feature maps at multiple depth levels and propagate them to the decoder via skip connections and concatenation~\cite{ronneberger2015u}. Instead, we alter the decoder such that, for every depth level $d$, after the two convolutional layers, an AAG first produces an attention map as the probabilistic prediction of a classifier (detailed below), then uses it to filter out activations from the input features map. 
Particularly, we use convolutional layers with $3\times3\times \mathscal{k}$ filters, being $\mathscal{k}$ the number of input channels, and produce the features map $\mathtensor{M}^{(d)}$. 
Then, the AAG classifier uses $\mathtensor{M}^{(d)}$ to predict a segmentation $\tilde{\mathimage{y}}^{(d)}$ at the given resolution level $d$. 
As a classifier, we use a convolutional layer with \mathscal{c} $1\times1\times \mathscal{k}$ filters (where \mathscal{c} is the number of possible classes, including the background). 
We do not apply any \emph{argmax} operation on its prediction, while we use a pixel-wise \emph{softmax} to give a probabilistic interpretation of the output: as a result, every pixel is associated to a probability of belonging to every considered class, which is important to have smoother gradients on the learned attention maps. 
We then slice the predicted array removing the channel associated to the background, and we use the multi-channel soft segmentation: 
i) as input to the discriminator at the same depth level; and 
ii) to produce an attention map, obtained by summing up the remaining channels into a 2D probabilistic map $\mathimage{a}^{(d)}$, localizing object positions in the image (Fig.~\ref{fig:architectures}).
To force the segmentor to use $\mathimage{a}^{(d)}$, we multiply the extracted features $\mathtensor{M}^{(d)}$ with $\mathimage{a}^{(d)}$ using the Hadamard product (gating process). The resulting features maps are upsampled to the next resolution level via a nearest-neighbor interpolation. After each convolutional layer, we use batch normalization~\cite{ioffe2015batch} and \textit{ReLU} activation function.

\paragraph*{Discriminator $\Delta(\cdot)$} 
We design an encoding architecture receiving \textit{real} or \textit{fake} inputs at multiple scales. This allows a multi-level interaction between $\Sigma(\cdot)$ and $\Delta(\cdot)$, and the \emph{direct} propagation of adversarial gradients into the AAGs. We refer to this multi-level interaction as \textit{Adversarial Deep Supervision} (ADS), as it regularises the output of AAG classifiers similarly to deep supervision, but using adversarial gradients (Fig.~\ref{fig:aag_components}).
The \textit{real samples} $\{\mathimage{y}^{(d)}\}_{d=1}^4$ consist of expert-made segmentations, that we supply at full or downsampled resolution at multiple discriminator depths, while \textit{fake samples} $\{\tilde{\mathimage{y}}^{(d)}\}_{d=1}^4$ are the multi-scale predictions of the segmentor. 
In both cases, the lower-resolution inputs ($d>1$) are supplied to the discriminator by simply concatenating them to the features maps it extracts at each depth $d$ (Fig.~\ref{fig:architectures}, right).

The discriminator is a convolutional encoder adapted from~\cite{chartsias2019disentangled}. At every depth $d$, at first, we process and downsample the features maps using a convolutional layer with $4\times 4 \times k$ kernels and stride of 2. The number of filters follows that of the segmentor encoder (e.g. 32, 64, 128, 256, 512). We also use spectral normalization~\cite{miyato2018spectral} to improve training. Obtained feature maps are then compressed with a second convolutional layer using 12 $1 \times 1 \times k$ filters. Both layers use \textit{tanh} activations. 

To improve the learning process and avoid overfitting, we make the adversarial game harder for the discriminator, using \textit{label noise}~\cite{salimans2016improved} and \textit{instance noise}~\cite{sonderby2016amortised}. In particular, we obtain label noise by a random flip of the discriminator labels (\textit{real} vs \textit{fake}) with a 10\% probability, while we apply instance noise as a Gaussian noise with zero mean and standard deviation of 0.2, that we add to the highest resolution input. 

Lastly, we compute the final prediction of the discriminator using a fully connected layer with scalar output ($\Delta(\mathimage{y})$, $\Delta(\tilde{\mathimage{y}})$).

\subsection{Loss Functions and Training Details}\label{subsec:loss_function_and_training}
We train the model minimizing supervised and adversarial objectives. In particular, we consider both contributions when scribble annotations are available for the input image, only the latter when we are using unlabeled data. 

\subsubsection*{Supervised Cost}
When scribbles are available, we train the segmentor to minimize a pixel-wise classification cost on the annotated pixels of the image-scribble pair $(\mathimage{x}, \mathimage{y_s})$, while, most importantly, we don't propagate any loss gradient trough the unlabeled pixels.
Crucially, we use the pixel-wise cross-entropy because it is shape-independent, and, to resolve the class imbalance problem, we multiply the per-class loss contribution by a scaling factor that accounts for the class cardinality.
We can write the supervised cost as:
\begin{align}\label{eq:l_sup}
    \mathcal{L}_{SUP}  
            = \mathds{1}(\mathimage{y_{s}}) * 
                        \big[ - \sum\nolimits_{\mathscal{i}=1}^{\mathscal{c}} 
                        \mathscal{w}_i \cdot 
                        \mathimage{y_s}_i \log(\tilde{\mathimage{y}}_i) \big],
\end{align}
where $i$ refers to each class and $\mathscal{c}$ is the number of classes. We choose the class scaling factor $\mathscal{w}_i = 1 - \mathscal{n}_i/\mathscal{n}_{tot}$, being $n_i$ the number of pixels with label $\mathscal{i}$ within $\mathimage{y_s}$, and $\mathscal{n}_{tot}$ the total number of annotated pixels.
To avoid loss contribution on unlabeled pixels, we multiply the result by the masking function $\mathds{1}(\mathimage{y_s})$, which returns 1 for annotated pixels, 0 otherwise. A similar formulation was suggested in~\cite{tang2018normalized} termed as Partial Cross-Entropy (PCE) loss but without the class balancing. Thus, we term our formulation as Weighted-PCE (WPCE).

\subsubsection*{Adversarial Cost} 
Adversarial objectives are the result of a minimax game~\cite{goodfellow2014generative} between segmentor and discriminator, where $\Delta(\cdot)$ is trained to maximize its capability of differentiating between real and generated segmentations, $\Sigma(\cdot)$ to predict segmentation masks that are good enough to trick the discriminator and minimize its performance.

To address the difficulties of training GANs, that can lead to training instability~\cite{mao2018effectiveness}, we adopt the Least Square GAN objective~\cite{mao2018effectiveness} which penalizes prediction errors of the discriminator based on their distances from the decision boundary.

Given an image $\mathimage{x}$ and an unpaired mask $\mathimage{y}$, we optimize $\Delta$ and $\Sigma$ according to: $\min_{\Delta} \mathcal{V}_{LS}(\Delta)$ and $\min_{\Sigma} \mathcal{V}_{LS}(\Sigma)$, where:
\begin{equation}\label{eq:lsgan}
\begin{split}
         & \mathcal{V}_{LS}(\Delta) = 
            \frac{1}{2} E_{\mathimage{y} \sim \mathdistrib{Y}}[(\Delta(\mathimage{y}) - 1)^2] 
            + 
            \frac{1}{2} E_{\mathimage{x} \sim \mathdistrib{X}}[(\Delta(\Sigma(\mathimage{x})) + 1)^2] 
        \\ &
        \mathcal{V}_{LS}(\Sigma) = 
            \frac{1}{2} E_{\mathimage{x} \sim \mathdistrib{X}}[(\Delta(\Sigma(\mathimage{x})) - 1)^2].
\end{split}
\end{equation}

\subsubsection*{Training Strategy}
We iterate the training of the model over two steps: i) optimization over a batch of weakly annotated images, and ii) optimization over a batch of unlabeled images.

When scribble annotations are available, we minimize $\mathcal{L} = \mathscal{a}_0 \mathcal{L}_{SUP} + \mathscal{a}_1 \mathcal{V}_{LS}(\Sigma)$. 
In particular, we compute $\mathscal{a}_0$ \emph{dynamically}, so that we don't need to tune it. We define: $\mathscal{a}_0= \frac{\norm{\mathcal{V}_{LS}(\Sigma)}}{\norm{\mathcal{L}_{SUP}}}$ to maintain a fixed ratio between the amplitude of supervised and adversarial costs throughout the entire training process, preventing one factor to prevail over the other.

When dealing with a batch of unlabeled images, we alternately optimize the model. First, we compute the discriminator loss, $\mathscal{a}_2\mathcal{V}_{LS}(\Delta)$, and update discriminator's weights to reduce it. Then, with the updated discriminator, we estimate the generator loss, $\mathscal{a}_3\mathcal{V}_{LS}(\Sigma)$, and optimize the generator's weights.

We give more importance to the supervised objective rather than the adversarial loss because the discriminator only evaluates if the predicted masks look realistic, while it does not say anything about their accuracy. Besides, the supervised cost requires the segmentor to learn the correct mapping from images to segmentation masks, which is what we are interested into.
Thus, we scale the adversarial contribution to be one order of magnitude smaller, setting $\mathscal{a}_1 = 0.1$ for training with weak supervision. Similarly, we use $\mathscal{a}_2 = \mathscal{a}_3 = 0.2$ to train generator and discriminator equally on the unlabeled data.%

We minimize the loss function using Adam~\cite{kingma2014adam} and a batch size of 12. Most importantly, learning from limited annotations can easily trap the model in sharp, bad, local minima because the training data poorly represents the actual data distribution. Thus, we promote the search of flat and more generalizable solutions using a cyclical learning rate~\cite{smith2017cyclical} with a period of 20 epochs, that we oscillate between $10^{-4}$ and $10^{-5}$. As a result, we observed a smoother loss function and more stable performance between subsequent epochs, diminishing the early stopping criterion effects (as also observed in~\cite{valvano2019temporal}). 
Similarly to previous work with weak annotations~\cite{lin2016scribblesup,dai2015boxsup}, we train the model until an early stopping criterion is met, and we arrest the training when the loss between predicted and real segmentations stops decreasing on a validation set. %

%
%
%
%

%
%
%

\section{Experimental Setup}
\subsection{Data}\label{sec:data}
Below, we first describe the adopted datasets; then, we detail the procedure used to generate scribble annotations; finally, we define how we construct train, validation, and test set.
We consider medical and vision datasets, for the segmentation of heart, abdominal organs, and human pose parts:
\begin{enumerate}
    \item \textbf{ACDC}~\cite{bernard2018deep}. 
    This dataset contains 2-dimensional cine-MR images obtained by 100 patients using various 1.5T and 3T MR scanners and different temporal resolutions. Manual segmentations are provided for the end-diastolic (ED) and end-systolic (ES) cardiac phases for right ventricle (RV), left ventricle (LV) and myocardium (MYO). We resample the data to 1.51$mm^2$ and cropped or padded them to match a size of $224\times224$. We normalize the images of each patient by removing the median and dividing by the interquartile range computed per volume. 
    \item \textbf{LVSC}~\cite{suinesiaputra2014collaborative}. 
    It contains gated SSFP cine images of 100 patients, obtained from a mix of 1.5T scanner types and imaging parameters. Manual segmentations are provided for the left ventricular myocardium (MYO) in all the cardiac phases. To compare with ACDC, we only consider segmentations for ES and ED phase instants. We resample images to the average resolution of 1.45$mm^2$ and crop or pad them to a size of $224\times224$. We normalize the images of each patient by removing the median and dividing by the interquartile range computed on his MRI scan.
    \item \textbf{CHAOS}~\cite{chaos}. It has abdominal MR images of 20 subjects, with segmentation masks of liver, kidneys, and spleen. We test our model on the T1 in-phase and T2 images. We resample images to a resolution of 1.89$mm^2$ and crop to $192\times192$ pixels, after normalising in $[-1, 1]$. 
    \item \textbf{PPSS}~\cite{luo2013pedestrian}. To demonstrate the broad utility of our method, we use the (non-medical) Pedestrian Parsing in Surveillance Scenes data. PPSS contains RGB images of pedestrians with occlusions, derived from 171 surveillance videos, using different cameras and resolutions. Besides images, ground truth segmentations are given for seven parts of the pedestrians: hair, face, upper clothes, arms, legs, shoes, and background. Since provided segmentations have size $80\times160$, we resample all the images to the same spatial resolution. We also normalize images between 0 and 1, dividing them by their maximum value. 
\end{enumerate}

\subsubsection*{Scribble Generation}\label{sec:data_scribble}
\begin{figure}[t]
    \centering
    \includegraphics[width=0.48\textwidth]{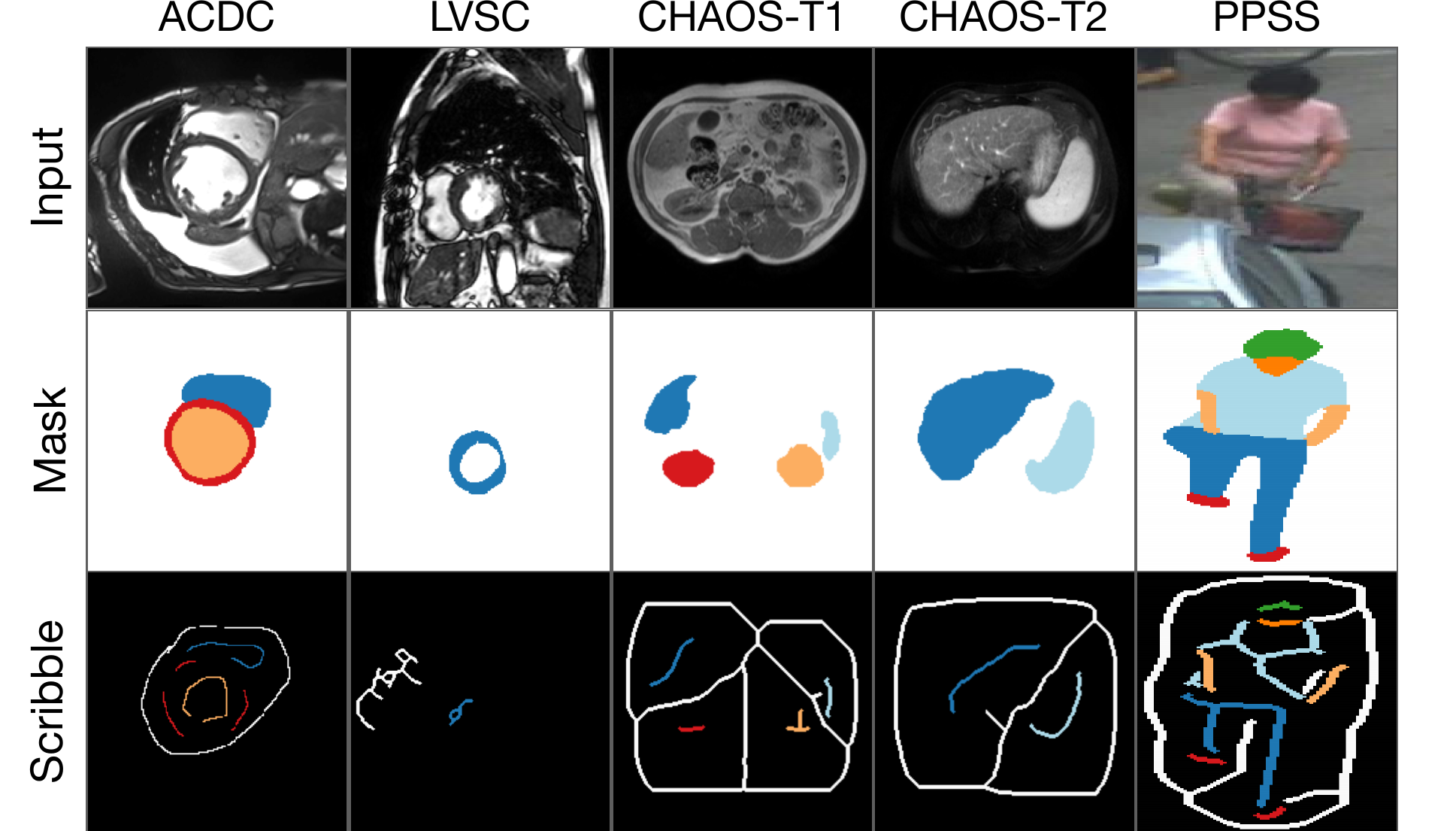}
    \caption{
    Example of scribbles for each dataset (images resized to the same resolution to easy visualisation). ACDC: manual annotations; LVSC: random walks; CHAOS and PPSS: skeletonisation. Please, refer to Section~\ref{sec:data_scribble} for additional details.
    }
    \label{fig:example_scribbles}
\end{figure}

To obtain scribbles with these datasets we follow different processes. Examples of those scribbles are shown in Fig.~\ref{fig:example_scribbles}.
Experts draw scribbles in a certain way (e.g., away from border regions). A dataset containing manual scribbles helps test a method more realistically than using simulated data from automatic procedures. Thus, in ACDC, we use ITK-SNAP~\cite{itksnap} to manually draw scribbles for ES and ED phases within the available segmentation masks. We obtained separate scribbles for RV, LV, and MYO, enabling us to test against ground truth segmentations. To identify pixels belonging to the background class (BGD), we draw an ulterior scribble approximately around the heart, while leaving the rest of the pixels unlabeled. Scribbles for RV, MYO, LV, BGD had an average (standard deviation) image coverage of 0.1 (0.1)\%, 0.2 (0.1)\%, 0.1 (0.1)\% and 10.4 (8.4)\%, respectively.

For CHAOS and PPSS, we obtained scribbles by eroding the available segmentation masks~\cite{rajchl2017employing}. For each object, we followed standard skeletonisation by iterative identification and removal of border pixels, until connectivity is lost. Resulting scribbles are deterministic, typically falling along the object's midline (as with manual ones~\cite{lin2016scribblesup}).

For LVSC, since MYO is thin, a skeleton is already too good of an approximation of the full mask. Thus, we generate scribbles with random walks. For every object, we first initialize an ``empty'' scribble, and define the 2D coordinates of a random pixel $\mathtensor{P}\equiv(\mathscal{x}_{\mathtensor{P}}, \mathscal{y}_{\mathtensor{P}})$ inside the segmentation mask. Then, we iterate 2500 times the steps: i) assign $\mathtensor{P}$ to the scribble; ii) randomly ``move'' in the image, adding or subtracting 1 to the coordinates of $\mathtensor{P}$; iii) if the new point belongs to the segmentation mask, assign the new coordinates to $\mathtensor{P}$. Scribbles for MYO and BGD had an average (standard deviation) image coverage of 0.2 (0.1) \% and 1.9 (0.5) \%, respectively.

\subsubsection*{Train, Validation, Test}\label{sec:data_splits}
We divided ACDC, LVSC, CHAOS-T1 and CHAOS-T2 datasets in groups of 70\%, 15\% and 15\% of patients for train, validation, and test set, respectively. Following seminal semi-supervised learning approaches~\cite{chartsias2019disentangled,salimans2016improved}, we additionally split the 70\% of training data into two halves, the first of which is used to train the segmentor $\Sigma(\cdot)$ with weak labels (image-scribble pairs), while we use \textit{only} the masks of the second half to train the discriminator $\Delta(\cdot)$. Correlations between groups are limited by: i) splitting the data by patient, rather than by images (limiting intra-subject leakage, as masks come from different subjects~\cite{chartsias2019disentangled}); and ii) discarding images associated to masks used to train the discriminator (thus, $\Sigma(\cdot)$ never sees images used to train $\Delta(\cdot)$).

For PPSS, following~\cite{luo2013pedestrian}, we use the video scenes from the last 71 cameras as test set, while we split images from the first 100 cameras to train (90\% of images) and validate (10\% of images) the model.
As with the medical datasets, we further divide the training volumes into two halves, and we use one of them to exclusively train the discriminator, using the segmentation masks and discarding the associated images. 

\subsection{Baseline, Benchmark Methods and Upper Bounds}
We evaluate the robustness of our method in terms of segmentation performance
compared with methods using different prior assumptions to regularise training with scribbles, summarized in Table~\ref{tab:prior}. 
In particular, we consider:
\begin{itemize}
    \item \textbf{UNet\textsubscript{PCE}} and \textbf{UNet\textsubscript{WPCE}}~\cite{tang2018normalized}:
    The UNet~\cite{ronneberger2015u} is one of the most common choices for training with fully annotated segmentation masks. We evaluate its behavior when trained with the PCE loss proposed for scribble supervision in~\cite{tang2018normalized}, or the WPCE loss introduced in (\ref{eq:l_sup}).
    
    \item \textbf{UNet\textsubscript{CRF}}:
    We also consider the previous UNet\textsubscript{WPCE} whose prediction is further processed by CRF as RNN layer~\cite{chen2017deeplab,zheng2015conditional},~\cite{monteiro2018conditional}. CRF as RNN models Conditional Random Fields as a recurrent neural network (RNN), incorporating the prior that nearby pixels with similar color intensities should be classified similarly in the segmentation mask. This layer can be trained end-to-end and does not require relabeling the training set. For ACDC and LVSC, we train such a layer with the same hyperparameters used for cardiac segmentation in~\cite{can2018learning}: $\sigma_\alpha = 160, \sigma_\beta = 3$ and $\sigma_\gamma = 10$. These parameters model the pairwise potentials of CRF as weighted Gaussians~\cite{zheng2015conditional}. 
    As in~\cite{can2018learning}, we use $5$ iterations for the RNN.
    For the other datasets, we set $\sigma_\gamma = 3$, as suggested in~\cite{zheng2015conditional}.
    
    \item \textbf{TS-UNet\textsubscript{CRF}}: 
    We compare our model to the two-steps procedure in~\cite{can2018learning}, using the variant modeling CRF as an RNN rather than a separate post-processing step, because no relevant difference was observed between the two, and this is simpler to use at inference. For the CRF as RNN, we used the same hyper-parameter setting of UNet\textsubscript{CRF}.
\end{itemize}
The above approaches do not exploit unpaired data during training. Thus, we also compare with two models that, despite not being proposed for weakly supervised learning, can exploit the extra unpaired data and learn data-driven shape priors:
\begin{itemize}
    \item \textbf{PostDAE}~\cite{larrazabal2020post}: this method trains a denoising autoencoder (DAE) on unpaired masks, and then uses it to post-processes the predictions of a pre-trained UNet. To train the UNet on scribbles and directly compare with our method, we use the WPCE loss.
    \item \textbf{UNet\textsubscript{D}}: 
    as in vanilla GANs, we train a UNet segmentor and a mask discriminator. The latter has the same architecture as ours (same capacity), but it receives inputs only at the highest resolution.
\end{itemize}
Lastly, we compare with the method of Zhang \textit{et al.}:
\begin{itemize}
    \item \textbf{ACCL}~\cite{zhang2020accl}: similar to UNet\textsubscript{D}, ACCL trains with scribbles using a PatchGAN discriminator~\cite{isola2017image}.
\end{itemize}
Finally, we consider two \textbf{upper bounds}, based on training with fully annotated segmentation masks: 
\begin{itemize}
    \item \textbf{UNet\textsuperscript{UB}}: 
    UNet trained with strong annotations. In this case, we train the UNet in a fully-supervised way using image-segmentation pairs and a weighted cross-entropy loss (with per-class weights defined as in (\ref{eq:l_sup})).
    \item \textbf{UNet\textsubscript{D}\textsuperscript{UB}}: 
    UNet as before, but with an additional vanilla mask discriminator, used to train on the unlabeled images. The discriminator is the same as that of our model, but it receives an input only at the highest resolution.
\end{itemize}

\begin{table}[t]
    \caption{Type of prior used by each model.}
    \label{tab:prior}
    \centering
    \begin{tabular}{l| c | c}
    		Model &
    		\begin{tabular}[c]{@{}c@{}}Uses\\Prior\end{tabular} 
    		& Type of Prior \\
    		\midrule
    		
    		UNet\textsubscript{PCE} & \xmark & $-$ \\
    		UNet\textsubscript{WPCE} & \xmark & $-$ \\
    		UNet\textsubscript{CRF} & \cmark & Mean Field Assumption~\cite{zheng2015conditional} \\
    		TS-UNet\textsubscript{CRF} & \cmark &  Mean Field Assumption~\cite{zheng2015conditional}\\
    		PostDAE & \cmark & Shape, via DAE \\
    		UNet\textsubscript{D} & \cmark & Shape, via Discriminator \\
    		ACCL & \cmark & Shape, via Patch Discriminator~\cite{isola2017image} \\
    		\hcell{Ours} & \hcell{\cmark} & \hcell{Multi-scale Shape, via AAGs} \\
    		\bottomrule
    	\end{tabular}
\end{table}

To compare methods, we always use same UNet segmentor, learning rate, batch size, and early stopping criterion. If a method does not use a discriminator, we simply discard the data we would have used to train $\Delta(\cdot)$. As Can \textit{et al.}~\cite{can2018learning}, we train the CRF as RNN layer of TS-UNet\textsubscript{CRF} with a learning rate $10^{4}$ times smaller than that used for the UNet training, and we update the RNN weights only every 10 iterations.

\subsubsection*{Evaluation} 
We measure performance with the multi-class Dice score: $Dice =\frac{2|\tilde{\mathimage{y}} \cdot \mathimage{y}|}{|\tilde{\mathimage{y}}|+|\mathimage{y}|}$,
where $\tilde{\mathimage{y}}$ and $\mathimage{y}$ are the multi-channel predicted and true segmentation, respectively.
To assess if improvements are statistically significant we use the non-parametric Wilcoxon test, and we denote statistical significance with $p \leq 0.05$ or $p \leq 0.01$ using one (*) or two (**) asterisks, respectively. We avoid multiple comparisons comparing our method only with the best benchmark model.

%
%
%
%
%
%
%
%
%


\section{Experiments and Discussion}
We present and discuss the performance of our method in various experimental scenarios. Our primary question is: \textit{Can scribbles replace per-pixel annotations} (Section~\ref{subsec:learning_from_scribbles},~\ref{subsec:segmentation_vs_masks}); \textit{and what happens when we have fewer scribble annotations, or less unpaired data} (Section~\ref{subsec:semi_sup_learning},~\ref{subsec:less_unpaired_data})? 
Then, we consider two natural questions that extend the applicability of our approach: 
\textit{Can we learn from multiple scribbles per training image} (Section~\ref{subsec:crowdsourcing})? 
\textit{Can we mix per-pixel annotations with scribbles during training} (Section~\ref{subsec:multitask})?
Finally, we ask: \textit{Why does Adversarial Attention Gating work} (Section~\ref{subsec:adversarial_deep_supervision})?

\begin{figure}[t]
    \centering
    \includegraphics[width=0.48\textwidth]{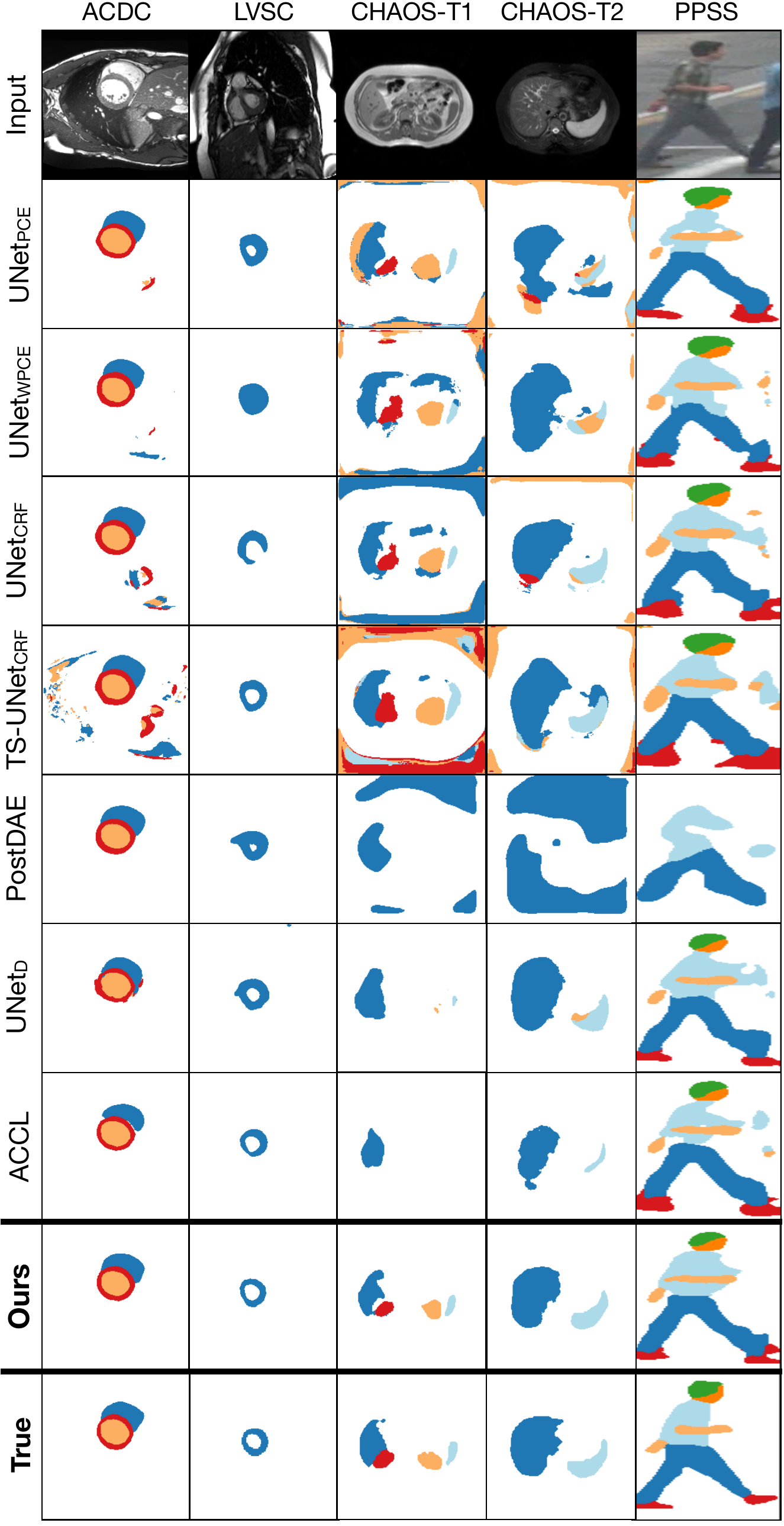}
    \caption{Example of predicted segmentation masks for the considered methods on each task. Observe that our approach (bottom row) learns spatial relationships in the image, thus preventing the prediction of isolated pixels in the mask, as well as unrealistic spatial relationship among the object parts.}
    \label{fig:panel_segmentations_dataset}
\end{figure}

\begin{table}[t]
    \caption{Dice average and standard deviation (subscript) obtained from each method on the test set, for medical and vision datasets. Leftmost column indicates if the learning algorithm has been trained with full mask or scribble annotations. The best method is in bold characters, while the second best is underlined; asterisks denote if their difference has statistical significance (* $p \leq 0.05$, ** $p \leq 0.01$).}
    \label{tab:performance_main}
    \centering
    \setlength{\tabcolsep}{1.5pt}
    
    \resizebox{\columnwidth}{!}{%
    \begin{tabular}{c| cl c|c|c|c|c}
    		\multicolumn{1}{l}{} & & & \multicolumn{5}{c}{Dataset}
    		\\ \cmidrule(lr){4-8}
    		        \multicolumn{1}{l}{} & & \multicolumn{1}{|l}{Model}
    		        & ACDC & LVSC & CHAOS-T1 & CHAOS-T2 & PPSS \\
    		\cmidrule[\arrayrulewidth]{2-8}
    		
    		\multirow{11}{*}{\rotatebox[origin=c]{90}{Supervision Type}} 
    		
    		& \multirow{8}{*}{\rotatebox[origin=c]{90}{Scribble}}  
    		& \multicolumn{1}{|l}{UNet\textsubscript{PCE}}   
    		                 & ~~79.0\textsubscript{06}     
    		                 & 62.3\textsubscript{09}     
    		                 & ~34.4\textsubscript{06}     
    		                 & 37.5\textsubscript{06}     
    		                 & ~~71.9\textsubscript{04} \\  
    		&& \multicolumn{1}{|l}{UNet\textsubscript{WPCE}}   
    		                 & ~~69.4\textsubscript{07}     
    		                 & 59.1\textsubscript{07}     
    		                 & ~40.0\textsubscript{05}
    		                 & \underline{52.1\textsubscript{05}}
    		                 & ~~69.3\textsubscript{04} \\
    		&& \multicolumn{1}{|l}{UNet\textsubscript{CRF}}
    		                 & ~~69.6\textsubscript{07} 
    		                 & 60.4\textsubscript{08}
    		                 & ~40.5\textsubscript{05}
    		                 & 44.7\textsubscript{06} 
    		                 & ~~68.8\textsubscript{04} \\
    		&& \multicolumn{1}{|l}{TS-UNet\textsubscript{CRF}}
    		                 & ~~37.3\textsubscript{08} 
    		                 & 50.5\textsubscript{07}
    		                 & ~29.3\textsubscript{05}
    		                 & 27.6\textsubscript{05} 
    		                 & ~~67.1\textsubscript{04} \\
    		&& \multicolumn{1}{|l}{PostDAE}
    		                 & ~~69.0\textsubscript{06} 
    		                 & 58.6\textsubscript{07}
    		                 & ~29.1\textsubscript{06}
    		                 & 35.5\textsubscript{05} 
    		                 & ~~67.5\textsubscript{04} \\
    		&& \multicolumn{1}{|l}{UNet\textsubscript{D}}
    		                 & ~~61.8\textsubscript{08} 
    		                 & 31.7\textsubscript{09}
    		                 & ~44.0\textsubscript{03}
    		                 & 46.3\textsubscript{01} 
    		                 & ~~73.1\textsubscript{04} \\
    		&& \multicolumn{1}{|l}{ACCL}
    		                 & ~~\underline{82.6\textsubscript{05}} 
    		                 & \textbf{65.9\textsubscript{08}}
    		                 & ~\underline{48.3\textsubscript{07}}
    		                 & 49.7\textsubscript{05}
    		                 & ~~\underline{73.2\textsubscript{04}} \\
    		&& \multicolumn{1}{|l}{\hcell{Ours}}   
    		                 & \hcell{**\textbf{84.3\textsubscript{04}}} 
    		                 & \hcell{\underline{65.5\textsubscript{08}}}
    		                 & \hcell{*\textbf{56.8\textsubscript{05}}} 
    		                 & \hcell{\textbf{57.8\textsubscript{04}}}
    		                 & \hcell{**\textbf{74.6\textsubscript{04}}} \\
    		
    		\cmidrule[\arrayrulewidth]{2-8}

    		& \multirow{2}{*}{\rotatebox[origin=c]{90}{Mask}} 
    		& \multicolumn{1}{|l}{UNet\textsuperscript{UB}}  
    		                 & ~~82.0\textsubscript{05}
    		                 & ~~67.2\textsubscript{07}
    		                 & ~60.8\textsubscript{06}
    		                 & 58.6\textsubscript{01}
    		                 & ~~72.8\textsubscript{04} \\
    		&& \multicolumn{1}{|l}{UNet\textsubscript{D}\textsuperscript{UB}}
    		                 & ~~83.9\textsubscript{05} 
    		                 & ~~67.9\textsubscript{09}
    		                 & ~63.9\textsubscript{05}
    		                 & 60.8\textsubscript{01}
    		                 & ~~77.2\textsubscript{04}  \\
    		\cmidrule[\heavyrulewidth]{2-8}
    	\end{tabular}
    }
\end{table}

\begin{figure*}
    \includegraphics[width=\textwidth]{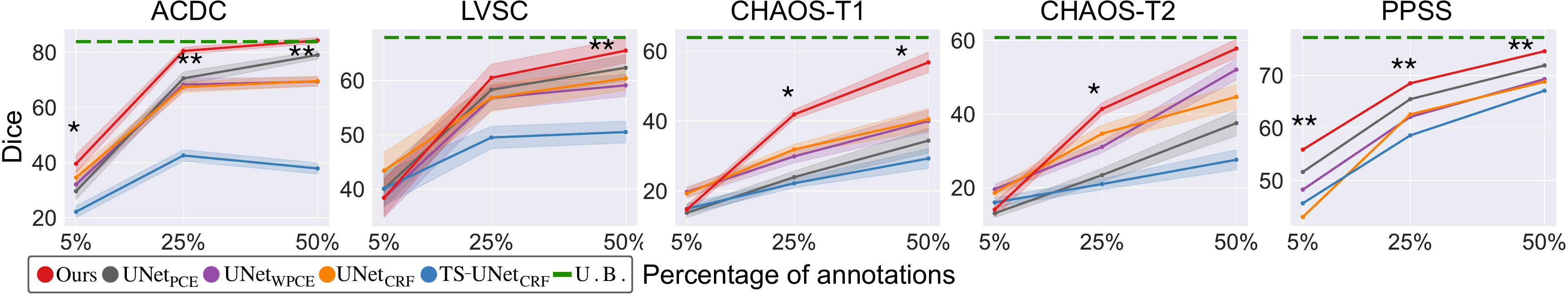}
	\caption{
	Dice score obtained on the test data by our and methods that don't use shape priors when changing the percentage of available labels in the training set (shaded bands show standard errors instead of deviation for clarity). As upper bound (U.B.) we consider UNet\textsubscript{D}\textsuperscript{UB}, trained using all the densely annotated masks. 
	Asterisks (*,**) have the same role as in Table~\ref{tab:performance_main}.
	}
	\label{fig:panel_segmentations_percentage}
\end{figure*}

\subsection{Learning from Scribbles}\label{subsec:learning_from_scribbles}
A prime contribution of our work is to close the performance gap between the most common strongly supervised models and weakly supervised approaches. Thus, we compare our method with other benchmarks and upper bounds quantitatively, in Table~\ref{tab:performance_main}, and qualitatively, in Fig.~\ref{fig:panel_segmentations_dataset}.

In particular, Table~\ref{tab:performance_main} reports average and standard deviation of the Dice score on test data for each dataset.\footnote{For ACDC, we also evaluated our model using the challenge server. After training our method on scribbles, we obtained an average (over the anatomical regions) Dice of 86.5\%. We report the full results in Section I of the Supplementary Material.} We clarify that, as discussed in Section~\ref{sec:data}, these results refer to training the segmentors with half of the annotated training images. We report Dice scores and the Hausdorff distances for each anatomical region of the medical datasets in Section I of the Supplementary Material.

Our method matches and sometimes even improves the performance of approaches trained only with strong supervision. As an example, we improve the Dice score of UNet\textsuperscript{UB} on both ACDC and PPSS. A result that further confirms the potential of weakly supervised approaches that use annotations which are much easier to collect than segmentation masks.

Moreover, as can be seen from the upper part of the table (methods trained with scribble supervision), we consistently improve segmentation results.\footnote{The only exception is on LVSC, where we have same results as ACCL} When compared to the $2^{nd}$ best model, we obtain up to $\sim$8.5\% of improvement on CHAOS-T1.
As our ablation study shows in Section~\ref{subsec:ablation}, such performance gains originate from the multi-scale interaction between adversarial signals and attention modules, which regularises the segmentor to predict both locally and globally consistent masks. In particular, our training strategy enforces multi-scale shape constraints, discouraging the appearance of isolated pixels and unrealistic spatial relationships between the object parts (Fig.~\ref{fig:panel_segmentations_dataset}).

Interestingly, we observe that weighting the loss contribution of each class based on their numerosity (UNet\textsubscript{PCE} vs UNet\textsubscript{WPCE}) is not always beneficial to the model, probably because, being sparse, scribble supervision suffers less than mask supervision from the class unbalance problem. However, when the class imbalance increases, e.g. with CHAOS-T1 and T2, weighting the PCE seems to be beneficial.
We also did not find evident performance boost in using CRF as RNN to post-process the UNet predictions (UNet\textsubscript{WPCE} vs UNet\textsubscript{CRF}). 

The two-step paradigm of TS-UNet\textsubscript{CRF} is one of the worst. We observed that errors reinforce themselves in self-learning schemes~\cite{chapelle2009semi}, and unreliable proposals in the relabeled training set lead the retrained model to fit to errors.\footnote{In this experiment, we explore the learning capability of the model and compare with benchmarks on the same ground. Thus, we did not enlarge scribbles as suggested by Can \textit{et al.}~\cite{can2018learning}~\cite{grady2006random}. 
With the enlarged scribbles, TS-UNet\textsubscript{CRF} improved from 37.3\% to 53.6\%, on ACDC. Doing the same for our method, gave no improvement (83.5\% vs 84.3\% from Table II). This illustrates that such additional training signal is useful for TS-UNet\textsubscript{CRF} but it is not necessary for our method. While we are not certain about the origins of this, we hypothesise that it is the adversarial discriminator that provides a similar training signal as those provided by the enlarged scribbles. 
}

Lastly, we discuss the performance of the methods that learn a shape prior from the unpaired masks. As Table~\ref{tab:performance_main} shows, post-processing the segmentor output with a DAE does not improve performance (PostDAE). As discussed by the PostDAE authors~\cite{larrazabal2020post}, a reason could be the poor performance of the segmentor which, when trained on scribbles, produces out-of-distribution segmentation masks for the DAE (i.e., the corrupted data used for training the DAE are not representative of the test-time segmentation errors). Sometimes, we even observed degenerate cases where the PostDAE always produces empty masks (CHAOS dataset and PPSS), or it completely omits some classes (ACDC). See Section III of the Supplementary Material, for visual examples of these and other models' failures.

Instead, mask discriminators are an effective choice (UNet\textsubscript{D} and ACCL). In fact, the discriminator can recover missing label information from the scribble-annotated data, and the model has competitive performance. However, our model generalises better across datasets.

\subsection{Segmentation Masks vs Scribbles}\label{subsec:segmentation_vs_masks}

To understand the trade-off between time-to-segment and type-of-annotations, we evaluate if it's better to collect many scribble annotations instead of few fully annotated images. Assuming that similar to bounding boxes~\cite{lin2014microsoft}, scribbles can be collected about $15\times$ faster than segmentations, annotating 35 images with scribbles on ACDC would require a similar time as two densely labelled masks. 
Some authors suggest the possibility to learn to segment using a few or even one single annotated sample~\cite{tajbakhsh2020embracing, shaban2017one, zhao2019data, chaitanya2019semi, liu2018learning, feyjie2020semi}. 
Thus, we want to compare the performance of our model using 35 scribble-annotations (Dice of 84.3\%) with that obtainable using two full masks and the Task-driven and Semi-supervised Data Augmentation (TSDA) method~\cite{chaitanya2019semi}.\footnote{We used the code provided by the authors at \url{https://github.com/krishnabits001/task_driven_data_augmentation}.} TSDA uses a GAN to learn realistic deformations and intensity transformations to apply on the annotated images and uses the augmented training set to optimise a UNet-like segmentor. 
We perform 3-fold cross-validation, using the same validation and test sets as before. We randomly selected two fully-annotated patients among the training subjects, and we learned the augmentation GAN with the unpaired images we assumed available (35 patients). 
With TSDA, we obtained an average Dice (standard deviation) of 56.8\% (13.5\%), which is considerably better than the standard training of a segmentor (Dice of 24.9\% (14.1\%)) but worse than other models trained with all the 35 scribble-annotated data (ACDC column, Table~\ref{tab:performance_main}). %

Our results confirm recent findings~\cite{asano2019critical} observing that despite a single image can be enough to train the first few layers of a CNN, deeper layers require additional labels.

Lastly, notice that TSDA data augmentation can be potentially integrated into our model, too.

\subsection{Model Robustness to Limited Annotations} \label{subsec:semi_sup_learning}
We analyze the robustness of the models with a scarcity of annotations in Fig.~\ref{fig:panel_segmentations_percentage}. In particular, we compare with methods that don't employ shape priors during training. In the experiments, we always use 50\% of training data to exclusively train the discriminator, if present in the method. The remaining 50\% is used to train the segmentor $\Sigma(\cdot)$, with varying amount of labels: e.g. ``5\%" means we train $\Sigma(\cdot)$ with 5\% of labeled and 45\% of unlabeled images (adversarial setup). As upper bound, we consider the results of UNet\textsubscript{D}\textsuperscript{UB}, trained with all the available image-segmentation pairs. 

As shown in Fig.~\ref{fig:panel_segmentations_percentage}, our model can rapidly approach the upper bound and, overall, it shows the best performance for almost every percentage of training annotations. 
With 5\% of weakly annotated data, our method performs slightly worse than other models in LVSC and CHAOS: however, the performance gap is not statistically significant.

\subsection{How Much Does the Model Rely on the Unpaired Data?}\label{subsec:less_unpaired_data}
Here, we investigate how much the model relies on the unpaired data by reducing the number of unpaired masks first, then the unpaired images.
In the first case, we trained the discriminator using only 5\% of the unpaired masks (3 ACDC patients) and the segmentor using all the scribbles. Despite training $\Delta(\cdot)$ with less masks, thanks to data augmentation (random roto-translations and instance noise), the model learned a robust shape prior and got a Dice of 83.7\% (5\%), i.e. less than 1\% decrease. Thus, the adversarial conditioning of the attention gates was still strong enough to correctly bias the segmentor to learn multi-scale relationships in the objects.\footnote{We conducted experiments also using more than 5\% of masks. Overall, we observed similar performance, with some fluctuations in Dice score due to the optimisation process. Such fluctuations originate from several factors: weight initialisation, training data, stochastic order of the batches presented to the network during training, etc. Minimising the performance gap between best- and worst-case scenario is a well-known problem of weakly supervised learning, and an active area of research~\cite{guo2019reliable}.} 
Secondly, we repeated the experiments in Section~\ref{subsec:semi_sup_learning} training our model without the additional unpaired images, and by varying the number of annotated data from 5\% to 50\%. At 5\% of annotations, we obtained an average (standard deviation in parenthesis) Dice of 22.5\% (10\%); with 25\% of scribble-annotations, a Dice of 75.0\% (8\%); and with 50\% of labels, we got 84.3\% (4\%). As can be seen, the model dependence on the number of unpaired images decreases when the number of scribble-annotated images (that are easy to collect) increases. 

Based on these experiments, we conclude that the model performs well even when the unpaired data are scarce, provided that enough scribble-annotations are available.

\begin{figure}
    \includegraphics[width=0.48\textwidth]{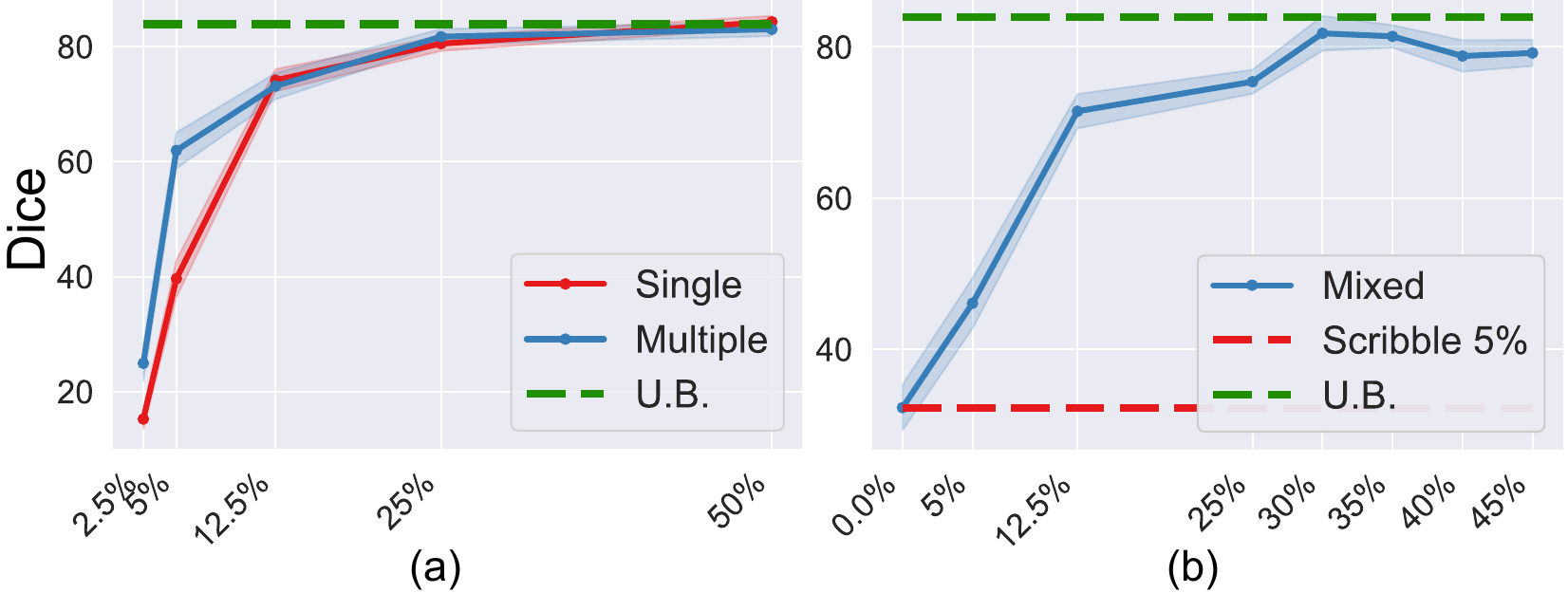}
    \caption{(a) Effect of training with labels from multiple annotators; and (b) performance in presence of mixed supervision (mask and scribbles) on ACDC. The upper bound (U.B.) is the UNet\textsubscript{D}\textsuperscript{UB}, trained with all the dense segmentation masks.}
    \label{fig:multiple_annotators_and_mixed_supervision}
\end{figure}

\subsection{Combining Multiple Scribbles: Simulating Crowdsourcing}\label{subsec:crowdsourcing}
Here we investigate the possibility to train our model using multiple scribbles per training image. This scenario simulates crowdsourcing applications, which are useful for annotating rare classes and to exploit different levels of expertise in annotators~\cite{lin2014microsoft,orting2019crowdsourcing}. 
We mimic the scribble annotations collected by three different ``sources'', using: i) expert-made scribbles; ii) scribbles approximated by segmentation masks skeletonization; iii) scribbles approximation by a random walk in the masks (see Section~\ref{sec:data} for a description of ii) and iii)). 

For every training image, we combine multiple scribbles summing up the supervised loss (\ref{eq:l_sup}) obtained for each of them: $\mathcal{L}_{SUP} = \sum_{i=1}^3 \mathcal{L}_{SUP}^i$. Thus, we consider multiple times pixels that are labeled across annotators, while considering `once' pixels labeled only from one annotator. Other ways of combining annotations are also possible (e.g., considering the union of the scribbles, or weighting differently each annotator~\cite{orting2019crowdsourcing}), but they are out of the scope of this manuscript. 

In Fig.~\ref{fig:multiple_annotators_and_mixed_supervision}a, we compare the Dice score of our method trained in a ``single'' vs a ``multiple'' annotator scenario. As can be seen, multiple scribbles have a regularising effect when the number of annotated data is scarce.

\begin{figure}[t]
    \centering
    \includegraphics[width=0.48\textwidth]{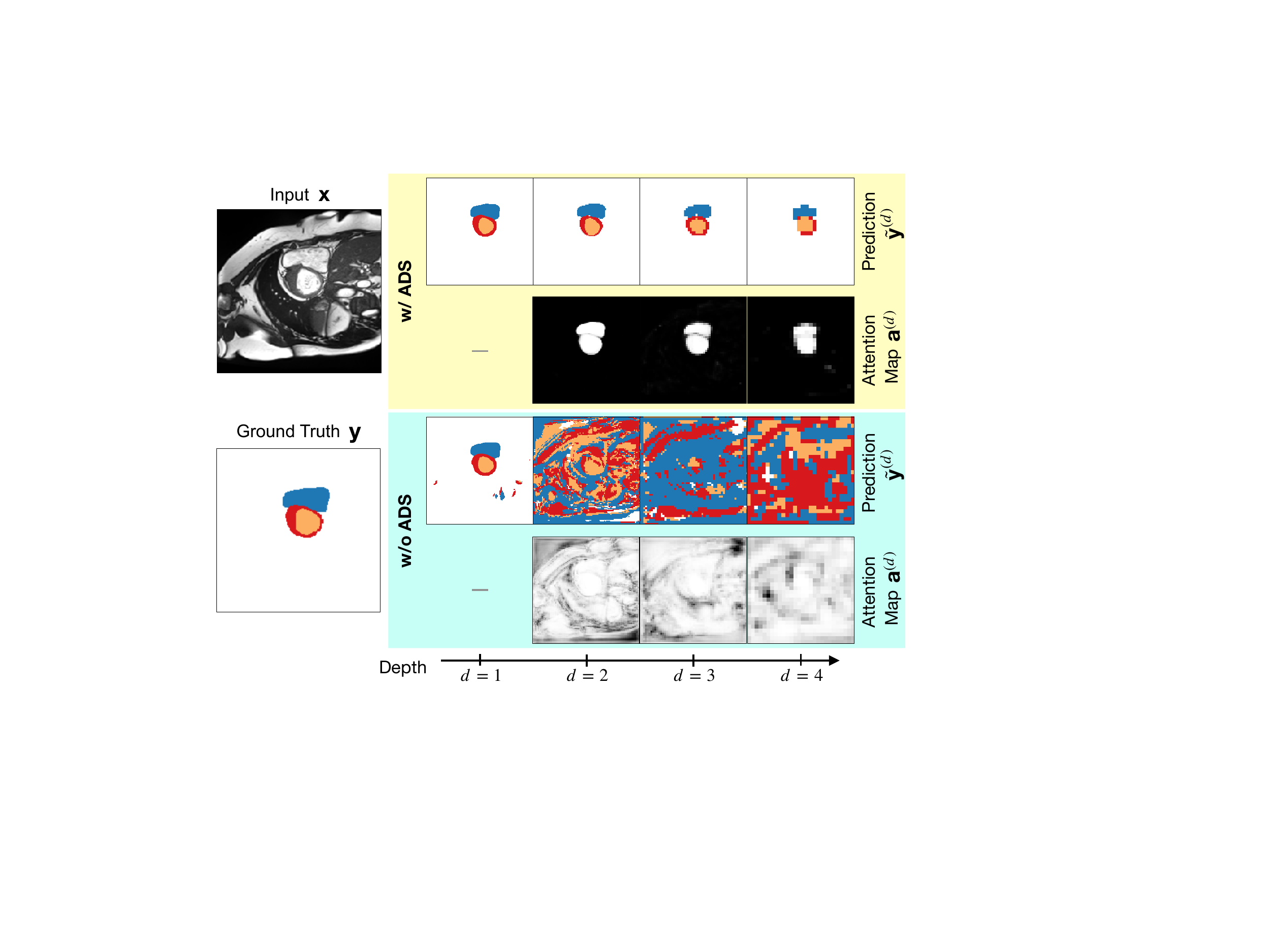}
    \caption{UNet-like segmentor with (top) vs without (bottom) adversarial conditioning of the attention gates in its decoder. Conditioned by an adversarial shape prior (w/ ADS), the model learns semantic attention maps able to localize the object to segment at multiple scales. Also, the shape prior encourages the segmentor to learn multi-scale relationships in the objects.}
    \label{fig:attention}
\end{figure}

\subsection{Multitask Learning: Combining Masks and Scribbles}\label{subsec:multitask}
Collecting homogeneous large-scale datasets can be difficult, but often we have access to multiple data sources, that can have different types of annotations. Here, we relax the assumption of using only scribble annotations, and investigate if we can train models that also leverage extra fully annotated data. 
For simplicity, we assume to have 5\% of scribble annotations, and we gradually introduce from 0\% to 45\% of fully-annotated images (for a maximum total of 50\% annotated data). 
We train the model using as loss: (\ref{eq:l_sup}) for scribble-annotated data, (\ref{eq:lsgan}) for unlabeled data, and the weighted cross-entropy for fully annotated images.
We report results on ACDC in Fig.~\ref{fig:multiple_annotators_and_mixed_supervision}b, showing that mixing scribble and mask supervision is feasible, and it can increase model performance. 
Although training only with masks is beyond the scope of this manuscript, we also investigated training in a fully supervised full mask setting. As expected, results show that training using only masks further improves segmentation performance (we report numbers in Section II of the Supplementary Material).

\subsection{Why does Adversarial Attention Gating work?}

\paragraph*{Prior-conditioned Attention Maps are Object Localizers}\label{subsec:attention_maps}
Here we show that, contrary to canonical attention gates, AAGs act as object localizers at multiple scales. In detail, we consider our attention mechanism with or without the adversarial conditioning (ADS). In both cases, the probability attention map is obtained as in Section~\ref{subsec:architectures}, and results from a $1 \times 1$ convolutional layer with softmax activation (that can be interpreted as a classifier), and a sum operation on all but one channel (see a summary in Fig.~\ref{fig:aag_components}). 
In Fig.~\ref{fig:attention} we illustrate: i) the most active channels in the classifier output, and ii) the predicted attention maps, at multiple depth levels $d$. 
As the attentions maps show (Fig.~\ref{fig:attention}, top), the adversarial conditioning of the attention gates encourages the segmentor at multiple scales to i) learn to localize objects of interest; and ii) suppress activations outside of them. 
Thus, scattered false positives (see UNet's prediction for $d=1$ in Fig.~\ref{fig:attention}) are prevented, and the model performance improves (see also Fig.~\ref{fig:panel_segmentations_dataset}).

\begin{figure}
    \centering
    \includegraphics[width=0.48\textwidth]{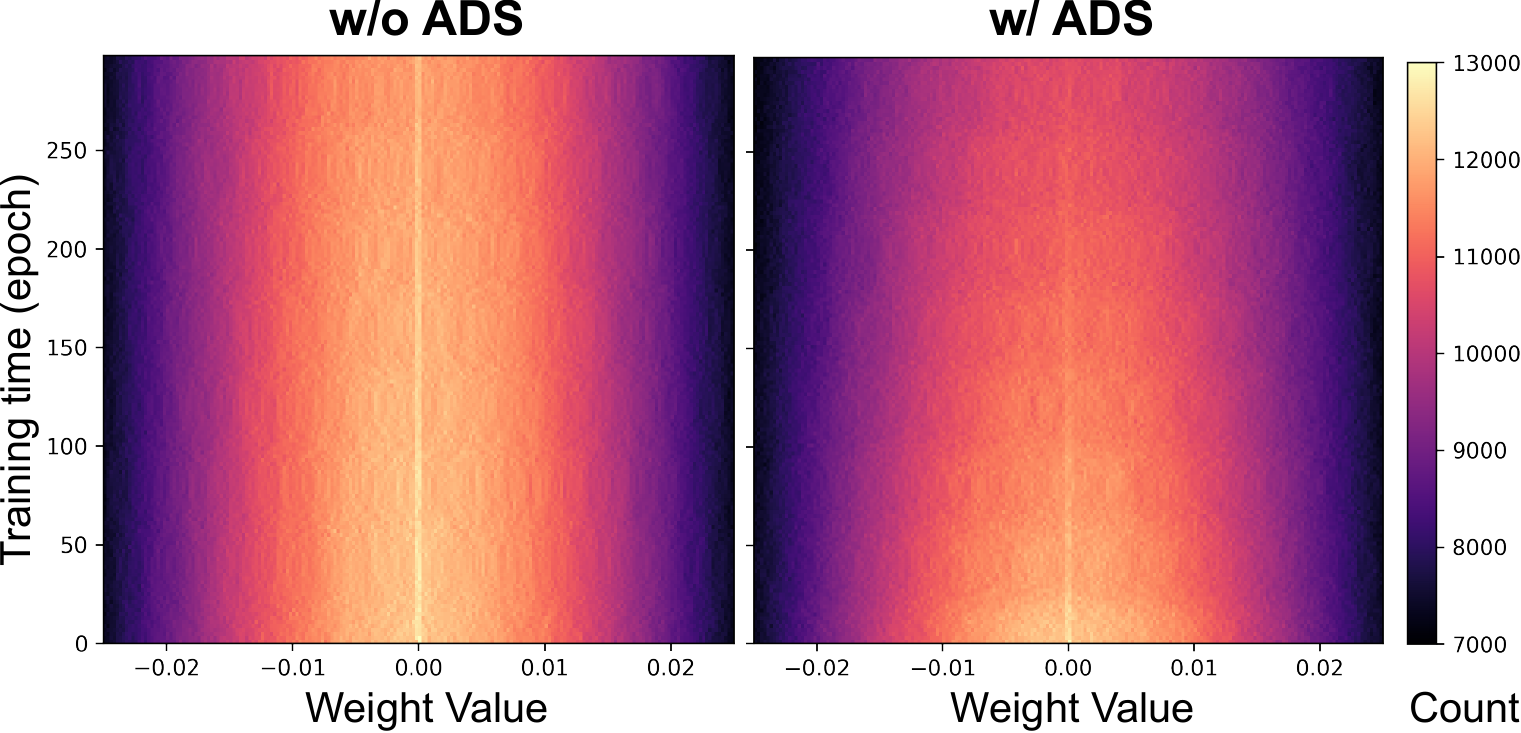}
    \caption{Weight distribution for the convolutional layers at depth d=4 of the segmentor. We compare how the weight distribution changes during training, with and without the use of ADS on the segmentor. Notice that ADS helps the layer training, and the initially narrow distribution becomes broader in time.}
    \label{fig:weight_distributions}
\end{figure}

\paragraph*{Adversarial Attention Gating Trains Deep Layers Better}\label{subsec:adversarial_deep_supervision}
We qualitatively show that AAGs increase the training of the segmentor deepest layers. In Fig.~\ref{fig:weight_distributions}, we show the distribution of weights values in the convolutional layers at depth $d=4$ in absence vs presence of adversarial conditioning (ADS) of the attention gates. 
As shown, attention gates with ADS force the segmentor to update its weights also in deeper layers, which would otherwise suffer from vanishing gradients~\cite{szegedy2015going,lee2015deeply}.

\begin{table}[t]
    \caption{Our ablations, as the name states, start with our model but remove: \#1: Only gating; \#2: Only ADS; \#3: Both Gating and ADS; \#4: Both ADS and the Discriminator; and finally \#5: ADS, the Discriminator and Gating.}
    \label{tab:ablation_acdc}
    \centering
    \resizebox{\columnwidth}{!}{%
        \begin{tabular}{l |c|c|c| ccc}
                \multirow{2}{*}{} 
                & \multirow{2}{*}{Attention}
                & \multicolumn{2}{c|}{Discriminator}
                & \multirow{2}{*}{5\%}
                & \multirow{2}{*}{25\%}
                & \multirow{2}{*}{50\%}\\
        		& & \begin{tabular}{@{}c@{}} single \end{tabular}
        		& \begin{tabular}{@{}c@{}} multi \end{tabular} &  &  & \\
        		\midrule
        		\hcell{Ours}
        		& \hcell{\checkmark} & \hcell{~} & \hcell{\checkmark}
        		& \hcell{40.7\textsubscript{09}}
        		& \hcell{80.6\textsubscript{06}}
        		& \hcell{84.3\textsubscript{05}} \\
        		
        		\#1 
        		& \checkmark & ~ & \checkmark
        		& 38.4\textsubscript{13}
        		& 79.1\textsubscript{06}
        		& 83.8\textsubscript{04} \\
        		
        		\#2 
        		& \checkmark & \checkmark & 
        		& 39.4\textsubscript{10} 
        		& 77.3\textsubscript{07}
        		& 84.0\textsubscript{05} \\
        		
        		\#3 
        		&  & \checkmark & 
        		& 55.8\textsubscript{10} 
        		& 60.2\textsubscript{07}
        		& 61.8\textsubscript{08} \\
        		
        		\#4 
        		& \checkmark &  & 
        		& 34.8\textsubscript{09} 
        		& 71.6\textsubscript{08} 
        		& 71.0\textsubscript{08} \\
        		
        		\#5 
        		&  &  & 
        		& 32.1\textsubscript{09} 
        		& 68.3\textsubscript{09}
        		& 69.4\textsubscript{07} \\
         		                
        		\bottomrule
        	\end{tabular}
        }
\end{table}

\paragraph*{Ablation Study}\label{subsec:ablation}
We show ablations on ACDC in Table~\ref{tab:ablation_acdc}. Removing ADS from the model, we leave the discriminator as a vanilla one, receiving inputs only at the highest resolution (classic GAN), while the segmentor remains unchanged. Unless otherwise stated, removing ADS we leave the attention gates in the segmentor, but without the adversarial conditioning (i.e. the segmentor is a UNet with classical self-attention; see Fig.~\ref{fig:aag_components}). 
When we completely remove the discriminator, the segmentor is trained just with scribble supervision and no adversarial signals. As  Table~\ref{tab:ablation_acdc} shows, each model component contributes to the final performance.

In particular, Table~\ref{tab:ablation_acdc} highlights that our model's success is not merely due to the use of additional unpaired images. In fact, if we compare with a classic GAN that also uses extra unpaired images, we Dice increases of 23\% when enough scribbles are available (compare ``Ours" vs ``\#3" at 25\% and 50\% of labels). %

From Table~\ref{tab:ablation_acdc}, we further observe that both ADS and the multiplicative gating are important aspects of the model, and they increase the segmentation quality of a similar amount (e.g., going from the ablation “\#3” to “\#2”, or to “\#1”, we obtain similar performance gains). This is not surprising: in fact, both the approaches enforce an attention process inside the segmentor. 
Specifically, the gating does so because it acts as an information bottleneck on what gets transmitted to the next convolutional block (i.e. it zeroes out unimportant information in the features maps). The ADS also enforces attention since it forces the segmentor to extract the information needed to predict realistic segmentations at every resolution. 
However, it is evident that ADS and the gating mechanism bring complementary advantages to the model, and it is when we combine \textit{both} of them that we reach the best results, at every percentage of labels (“Ours” vs “\#2”, “Ours” vs “\#3”).

Finally, we compared the use of the PCE vs WPCE loss to train the full model. With PCE, we obtained a Dice of: 25.2 (11), 74.0 (7), 83.4 (5) for 5\%, 25\% and 50\% of labels, respectively. With WPCE, our method performs better. We believe that this happens because PCE is intrinsically biased to penalize more the errors of the class having more annotated pixels. On the contrary, the WPCE loss is invariant to the number of annotated pixels. Thus, with WPCE, the discriminator can more easily bias the segmentor to predict masks which reflect the expected ratio between the organs/parts sizes and make them look realistic, ultimately improving segmentation performance.


\section{Conclusion}
We introduce a novel strategy to learn object segmentation using scribble supervision and a learned multi-scale shape prior. In an adversarial game, we force a segmentor to predict masks that satisfy short- and long-range dependencies in the image, narrowing down or eliminating the performance gap from strongly supervised models on medical and non-medical datasets. Fundamental to the success of our method are the proposed generalization of deep supervision and the novel adversarial conditioning of attention modules in the segmentor.

We show the robustness of our approach in diverse training scenarios, including: a varying number of scribble annotations in the training set, multiple annotators for an image (crowdsourcing), and the possibility to include fully annotated images during training. In the future, it would be interesting to explore the introduction of other types of multi-scale shape priors, such as those obtained by multi-scale VAEs, which can take into account also segmentation uncertainty.
Furthermore, it would be exciting to study other variants of the proposed attention gates, without relying on multiplicative gating operations and thus on background/foreground object segmentation tasks. It would also be interesting to explore the application of these gates for other tasks which could benefit from multi-scale adversarial signals, such as image registration~\cite{krebs2019learning}, conditional image generation~\cite{azadi2019semantic} and localised style transfer~\cite{kurzman2019class}.

Hoping to inspire new studies in weakly-supervised learning, we release manual scribble annotations for ACDC data, and the code used for the experiments.

\ifCLASSOPTIONcaptionsoff
  \newpage
\fi

\bibliographystyle{IEEEtran}
\bibliography{references}

\begin{thebibliography}{10}
\providecommand{\url}[1]{#1}
\csname url@samestyle\endcsname
\providecommand{\newblock}{\relax}
\providecommand{\bibinfo}[2]{#2}
\providecommand{\BIBentrySTDinterwordspacing}{\spaceskip=0pt\relax}
\providecommand{\BIBentryALTinterwordstretchfactor}{4}
\providecommand{\BIBentryALTinterwordspacing}{\spaceskip=\fontdimen2\font plus
\BIBentryALTinterwordstretchfactor\fontdimen3\font minus
  \fontdimen4\font\relax}
\providecommand{\BIBforeignlanguage}[2]{{%
\expandafter\ifx\csname l@#1\endcsname\relax
\typeout{** WARNING: IEEEtran.bst: No hyphenation pattern has been}%
\typeout{** loaded for the language `#1'. Using the pattern for}%
\typeout{** the default language instead.}%
\else
\language=\csname l@#1\endcsname
\fi
#2}}
\providecommand{\BIBdecl}{\relax}
\BIBdecl

\bibitem{chapelle2009semi}
O.~Chapelle, B.~Scholkopf, and A.~Zien, ``{Semi-supervised learning},''
  \emph{IEEE Trans Neural Networks}, vol.~20, no.~3, pp. 542--542, 2009.

\bibitem{cheplygina2019not}
V.~Cheplygina, M.~de~Bruijne, and J.~P. Pluim, ``Not-so-supervised: a survey of
  semi-supervised, multi-instance, and transfer learning in medical image
  analysis,'' \emph{MIA}, vol.~54, pp. 280--296, 2019.

\bibitem{zhou2019prior}
Y.~Zhou, Z.~Li, S.~Bai, C.~Wang, X.~Chen, M.~Han, E.~Fishman, and A.~L. Yuille,
  ``Prior-aware neural network for partially-supervised multi-organ
  segmentation,'' in \emph{ICCV}, 2019, pp. 10\,672--10\,681.

\bibitem{khoreva2017simple}
A.~Khoreva, R.~Benenson, J.~Hosang, M.~Hein, and B.~Schiele, ``Simple does it:
  Weakly supervised instance and semantic segmentation,'' in \emph{CVPR}, 2017,
  pp. 876--885.

\bibitem{can2018learning}
Y.~B. Can, K.~Chaitanya, B.~Mustafa, L.~M. Koch, E.~Konukoglu, and C.~F.
  Baumgartner, ``Learning to segment medical images with scribble-supervision
  alone,'' in \emph{DLMIA/ML-CDS}.\hskip 1em plus 0.5em minus 0.4em\relax
  Springer, 2018, pp. 236--244.

\bibitem{souly2017semi}
N.~Souly, C.~Spampinato, and M.~Shah, ``Semi supervised semantic segmentation
  using generative adversarial network,'' in \emph{ICCV}, 2017, pp. 5688--5696.

\bibitem{tajbakhsh2020embracing}
N.~Tajbakhsh, L.~Jeyaseelan, Q.~Li, J.~N. Chiang, Z.~Wu, and X.~Ding,
  ``{Embracing imperfect datasets: A review of deep learning solutions for
  medical image segmentation},'' \emph{MIA}, p. 101693, 2020.

\bibitem{lin2014microsoft}
T.-Y. Lin, M.~Maire, S.~Belongie, J.~Hays, P.~Perona, D.~Ramanan,
  P.~Doll{\'a}r, and C.~L. Zitnick, ``{Microsoft COCO: Common Objects in
  Context},'' in \emph{ECCV}.\hskip 1em plus 0.5em minus 0.4em\relax Springer,
  2014, pp. 740--755.

\bibitem{goodfellow2014generative}
I.~Goodfellow, J.~Pouget-Abadie, M.~Mirza, B.~Xu, D.~Warde-Farley, S.~Ozair,
  A.~Courville, and Y.~Bengio, ``Generative adversarial nets,'' in
  \emph{NeurIPS}, 2014, pp. 2672--2680.

\bibitem{denton2015deep}
E.~L. Denton, S.~Chintala, R.~Fergus \emph{et~al.}, ``Deep generative image
  models using a laplacian pyramid of adversarial networks,'' in
  \emph{NeurIPS}, 2015, pp. 1486--1494.

\bibitem{karras2017progressive}
T.~Karras, T.~Aila, S.~Laine, and J.~Lehtinen, ``Progressive growing of gans
  for improved quality, stability, and variation,'' \emph{ICLR}, 2017.

\bibitem{luo2018macro}
Y.~Luo, Z.~Zheng, L.~Zheng, T.~Guan, J.~Yu, and Y.~Yang, ``Macro-micro
  adversarial network for human parsing,'' in \emph{ECCV}, 2018, pp. 418--434.

\bibitem{schlemper2019attention}
J.~Schlemper, O.~Oktay, M.~Schaap, M.~Heinrich, B.~Kainz, B.~Glocker, and
  D.~Rueckert, ``Attention gated networks: Learning to leverage salient regions
  in medical images,'' \emph{MIA}, vol.~53, pp. 197--207, 2019.

\bibitem{larrazabal2020post}
A.~J. Larrazabal, C.~Mart{\'\i}nez, B.~Glocker, and E.~Ferrante, ``{Post-DAE:
  Anatomically plausible segmentation via post-processing with Denoising
  Autoencoders},'' \emph{IEEE TMI}, 2020.

\bibitem{painchaud2020cardiac}
N.~Painchaud, Y.~Skandarani, T.~Judge, O.~Bernard, A.~Lalande, and P.-M.
  Jodoin, ``Cardiac segmentation with strong anatomical guarantees,''
  \emph{IEEE Transactions on Medical Imaging}, vol.~39, no.~11, pp. 3703--3713,
  2020.

\bibitem{bernard2018deep}
O.~Bernard, A.~Lalande, C.~Zotti, F.~Cervenansky, X.~Yang, P.-A. Heng,
  I.~Cetin, K.~Lekadir, Camara \emph{et~al.}, ``{Deep learning techniques for
  automatic {MRI} cardiac multi-structures segmentation and diagnosis: Is the
  problem solved?}'' \emph{IEEE TMI}, vol.~37, no.~11, pp. 2514--2525, 2018.

\bibitem{suinesiaputra2014collaborative}
A.~Suinesiaputra, B.~R. Cowan, A.~O. Al-Agamy, M.~A. Elattar, N.~Ayache, A.~S.
  Fahmy, A.~M. Khalifa, P.~Medrano-Gracia, M.-P. Jolly, A.~H. Kadish
  \emph{et~al.}, ``A collaborative resource to build consensus for automated
  left ventricular segmentation of cardiac mr images,'' \emph{MIA}, vol.~18,
  no.~1, pp. 50--62, 2014.

\bibitem{chaos}
A.~Emre~Kavur, N.~Sinem~Gezer, M.~Bar{\i}ș, P.-H. Conze, V.~Groza,
  D.~Duy~Pham, S.~Chatterjee, P.~Ernst, S.~{\"O}zkan, B.~Baydar \emph{et~al.},
  ``{CHAOS Challenge--Combined (CT-MR) Healthy Abdominal Organ Segmentation},''
  \emph{arXiv}, pp. arXiv--2001, 2020.

\bibitem{luo2013pedestrian}
P.~Luo, X.~Wang, and X.~Tang, ``Pedestrian parsing via deep decompositional
  network,'' in \emph{ICCV}, 2013, pp. 2648--2655.

\bibitem{lin2016scribblesup}
D.~Lin, J.~Dai, J.~Jia, K.~He, and J.~Sun, ``{Scribblesup: Scribble-supervised
  convolutional networks for semantic segmentation},'' in \emph{CVPR}, 2016,
  pp. 3159--3167.

\bibitem{ji2019scribble}
Z.~Ji, Y.~Shen, C.~Ma, and M.~Gao, ``Scribble-based hierarchical weakly
  supervised learning for brain tumor segmentation,'' in \emph{MICCAI}.\hskip
  1em plus 0.5em minus 0.4em\relax Springer, 2019, pp. 175--183.

\bibitem{tang2018regularized}
M.~Tang, F.~Perazzi, A.~Djelouah, I.~Ben~Ayed, C.~Schroers, and Y.~Boykov,
  ``{On regularized losses for weakly-supervised CNN segmentation},'' in
  \emph{ECCV}, 2018, pp. 507--522.

\bibitem{chen2017deeplab}
L.-C. Chen, G.~Papandreou, I.~Kokkinos, K.~Murphy, and A.~L. Yuille,
  ``{Deeplab: Semantic image segmentation with deep convolutional nets, atrous
  convolution, and fully connected CRFs},'' \emph{IEEE PAMI}, vol.~40, no.~4,
  pp. 834--848, 2017.

\bibitem{zheng2015conditional}
S.~Zheng, S.~Jayasumana, B.~Romera-Paredes, V.~Vineet, Z.~Su, D.~Du, C.~Huang,
  and P.~H. Torr, ``{Conditional random fields as recurrent neural networks},''
  in \emph{ICCV}, 2015, pp. 1529--1537.

\bibitem{zhang2020accl}
P.~Zhang, Y.~Zhong, and X.~Li, ``{ACCL: Adversarial constrained-CNN loss for
  weakly supervised medical image segmentation},'' \emph{arXiv:2005.00328},
  2020.

\bibitem{isola2017image}
P.~Isola, J.-Y. Zhu, T.~Zhou, and A.~A. Efros, ``Image-to-image translation
  with conditional adversarial networks,'' in \emph{CVPR}, 2017, pp.
  1125--1134.

\bibitem{nosrati2016incorporating}
M.~S. Nosrati and G.~Hamarneh, ``Incorporating prior knowledge in medical image
  segmentation: a survey,'' \emph{arXiv:1607.01092}, 2016.

\bibitem{clough2019topological}
J.~R. Clough, I.~Oksuz, N.~Byrne, V.~A. Zimmer, J.~A. Schnabel, and A.~P. King,
  ``{A Topological Loss Function for Deep-Learning based Image Segmentation
  using Persistent Homology},'' \emph{arXiv:1910.01877}, 2019.

\bibitem{oktay2017anatomically}
O.~Oktay, E.~Ferrante, K.~Kamnitsas, M.~Heinrich, W.~Bai, J.~Caballero, S.~A.
  Cook, A.~De~Marvao, T.~Dawes, D.~P. O‘Regan \emph{et~al.}, ``{Anatomically
  constrained neural networks (ACNNs): application to cardiac image enhancement
  and segmentation},'' \emph{IEEE TMI}, vol.~37, no.~2, pp. 384--395, 2017.

\bibitem{dalca2018anatomical}
A.~V. Dalca, J.~Guttag, and M.~R. Sabuncu, ``{Anatomical priors in
  convolutional networks for unsupervised biomedical segmentation},'' in
  \emph{CVPR}, 2018, pp. 9290--9299.

\bibitem{kingma2014auto}
D.~P. Kingma and M.~Welling, ``Auto-encoding variational bayes,'' \emph{ICLR},
  2014.

\bibitem{yue2019cardiac}
Q.~Yue, X.~Luo, Q.~Ye, L.~Xu, and X.~Zhuang, ``{Cardiac segmentation from LGE
  MRI using deep neural network incorporating shape and spatial priors},'' in
  \emph{MICCAI}.\hskip 1em plus 0.5em minus 0.4em\relax Springer, 2019, pp.
  559--567.

\bibitem{painchaud2019cardiac}
N.~Painchaud, Y.~Skandarani, T.~Judge, O.~Bernard, A.~Lalande, and P.-M.
  Jodoin, ``{Cardiac MRI segmentation with strong anatomical guarantees},'' in
  \emph{MICCAI}.\hskip 1em plus 0.5em minus 0.4em\relax Springer, 2019, pp.
  632--640.

\bibitem{kervadec2019constrained}
H.~Kervadec, J.~Dolz, M.~Tang, E.~Granger, Y.~Boykov, and I.~B. Ayed,
  ``{Constrained-CNN losses for weakly supervised segmentation},'' \emph{MIA},
  vol.~54, pp. 88--99, 2019.

\bibitem{dalca2019unsupervised}
A.~V. Dalca, E.~Yu, P.~Golland, B.~Fischl, M.~R. Sabuncu, and J.~E. Iglesias,
  ``{Unsupervised deep learning for Bayesian brain MRI segmentation},'' in
  \emph{MICCAI}.\hskip 1em plus 0.5em minus 0.4em\relax Springer, 2019, pp.
  356--365.

\bibitem{kohl2018probabilistic}
S.~Kohl, B.~Romera-Paredes, C.~Meyer, J.~De~Fauw, J.~R. Ledsam, K.~Maier-Hein,
  S.~A. Eslami, D.~J. Rezende, and O.~Ronneberger, ``{A probabilistic U-Net for
  segmentation of ambiguous images},'' in \emph{NeurIPS}, 2018, pp. 6965--6975.

\bibitem{baumgartner2019phiseg}
C.~F. Baumgartner, K.~C. Tezcan, K.~Chaitanya, A.~M. H{\"o}tker, U.~J.
  Muehlematter, K.~Schawkat, A.~S. Becker, O.~Donati, and E.~Konukoglu,
  ``{Phiseg: Capturing uncertainty in medical image segmentation},'' in
  \emph{MICCAI}.\hskip 1em plus 0.5em minus 0.4em\relax Springer, 2019, pp.
  119--127.

\bibitem{chartsias2019disentangled}
A.~Chartsias, T.~Joyce, G.~Papanastasiou, S.~Semple, M.~Williams, D.~E. Newby,
  R.~Dharmakumar, and S.~A. Tsaftaris, ``Disentangled representation learning
  in cardiac image analysis,'' \emph{Medical image analysis}, vol.~58, p.
  101535, 2019.

\bibitem{yang2019unsupervised}
J.~Yang, N.~C. Dvornek, F.~Zhang, J.~Chapiro, M.~Lin, and J.~S. Duncan,
  ``Unsupervised domain adaptation via disentangled representations:
  Application to cross-modality liver segmentation,'' in \emph{MICCAI}.\hskip
  1em plus 0.5em minus 0.4em\relax Springer, 2019, pp. 255--263.

\bibitem{valvano2019temporal}
G.~Valvano, A.~Chartsias, A.~Leo, and S.~A. Tsaftaris, ``Temporal consistency
  objectives regularize the learning of disentangled representations,'' in
  \emph{DART}.\hskip 1em plus 0.5em minus 0.4em\relax Springer, 2019, pp.
  11--19.

\bibitem{xue2018segan}
Y.~Xue, T.~Xu, H.~Zhang, L.~R. Long, and X.~Huang, ``{SegAN: Adversarial
  network with multi-scale l1 loss for medical image segmentation},''
  \emph{Neuroinformatics}, vol.~16, no. 3-4, pp. 383--392, 2018.

\bibitem{vaswani2017attention}
A.~Vaswani, N.~Shazeer, N.~Parmar, J.~Uszkoreit, L.~Jones, A.~N. Gomez,
  {\L}.~Kaiser, and I.~Polosukhin, ``Attention is all you need,'' in
  \emph{NeurIPS}, 2017, pp. 5998--6008.

\bibitem{Jetley2018}
S.~Jetley, N.~A. Lord, N.~Lee, and P.~H.~S. Torr, ``{Learn To Pay Attention},''
  \emph{ICLR}, 2018.

\bibitem{oktay2018attention}
O.~Oktay, J.~Schlemper, L.~L. Folgoc, M.~Lee, M.~Heinrich, K.~Misawa, K.~Mori,
  S.~McDonagh, N.~Y. Hammerla, B.~Kainz \emph{et~al.}, ``{Attention U-net:
  Learning where to look for the pancreas},'' \emph{MIDL}, 2018.

\bibitem{wang2018deep}
Y.~Wang, Z.~Deng, X.~Hu, L.~Zhu, X.~Yang, X.~Xu, P.-A. Heng, and D.~Ni, ``Deep
  attentional features for prostate segmentation in ultrasound,'' in
  \emph{MICCAI}.\hskip 1em plus 0.5em minus 0.4em\relax Springer, 2018, pp.
  523--530.

\bibitem{sinha2020multi}
A.~Sinha and J.~Dolz, ``Multi-scale self-guided attention for medical image
  segmentation,'' \emph{IEEE J Biomed Health Inform}, 2020.

\bibitem{fu2019dual}
J.~Fu, J.~Liu, H.~Tian, Y.~Li, Y.~Bao, Z.~Fang, and H.~Lu, ``Dual attention
  network for scene segmentation,'' in \emph{CVPR}, 2019, pp. 3146--3154.

\bibitem{vahdat2020nvae}
A.~Vahdat and J.~Kautz, ``{NVAE: A Deep Hierarchical Variational
  Autoencoder},'' \emph{arXiv:2007.03898}, 2020.

\bibitem{ronneberger2015u}
O.~Ronneberger, P.~Fischer, and T.~Brox, ``U-net: Convolutional networks for
  biomedical image segmentation,'' in \emph{MICCAI}.\hskip 1em plus 0.5em minus
  0.4em\relax Springer, 2015, pp. 234--241.

\bibitem{ioffe2015batch}
S.~Ioffe and C.~Szegedy, ``{Batch normalization: Accelerating deep network
  training by reducing internal covariate shift},'' \emph{ICML}, 2015.

\bibitem{miyato2018spectral}
T.~Miyato, T.~Kataoka, M.~Koyama, and Y.~Yoshida, ``Spectral normalization for
  generative adversarial networks,'' \emph{ICLR}, 2018.

\bibitem{salimans2016improved}
T.~Salimans, I.~Goodfellow, W.~Zaremba, V.~Cheung, A.~Radford, and X.~Chen,
  ``Improved techniques for training {GANs},'' in \emph{NeurIPS}, 2016, pp.
  2234--2242.

\bibitem{sonderby2016amortised}
C.~K. S{\o}nderby, J.~Caballero, L.~Theis, W.~Shi, and F.~Husz{\'a}r,
  ``Amortised map inference for image super-resolution,'' \emph{ICLR}, 2017.

\bibitem{tang2018normalized}
M.~Tang, A.~Djelouah, F.~Perazzi, Y.~Boykov, and C.~Schroers, ``{Normalized cut
  loss for weakly-supervised CNN segmentation},'' in \emph{CVPR}, 2018, pp.
  1818--1827.

\bibitem{mao2018effectiveness}
X.~Mao, Q.~Li, H.~Xie, R.~Y.~K. Lau, Z.~Wang, and S.~P. Smolley, ``On the
  effectiveness of least squares generative adversarial networks,'' \emph{IEEE
  PAMI}, 2018.

\bibitem{kingma2014adam}
D.~P. Kingma and J.~Ba, ``Adam: A method for stochastic optimization,''
  \emph{ICLR}, 2015.

\bibitem{smith2017cyclical}
L.~N. Smith, ``Cyclical learning rates for training neural networks,'' in
  \emph{WACV}.\hskip 1em plus 0.5em minus 0.4em\relax IEEE, 2017, pp. 464--472.

\bibitem{dai2015boxsup}
J.~Dai, K.~He, and J.~Sun, ``{Boxsup: Exploiting bounding boxes to supervise
  convolutional networks for semantic segmentation},'' in \emph{ICCV}, 2015,
  pp. 1635--1643.

\bibitem{itksnap}
P.~A. Yushkevich, J.~Piven, H.~Cody~Hazlett, R.~Gimpel~Smith, S.~Ho, J.~C. Gee,
  and G.~Gerig, ``{User-Guided 3D Active Contour Segmentation of Anatomical
  Structures: Significantly Improved Efficiency and Reliability},''
  \emph{Neuroimage}, vol.~31, no.~3, pp. 1116--1128, 2006.

\bibitem{rajchl2017employing}
M.~Rajchl, L.~M. Koch, C.~Ledig, J.~Passerat-Palmbach, K.~Misawa, K.~Mori, and
  D.~Rueckert, ``{Employing weak annotations for medical image analysis
  problems},'' \emph{arXiv:1708.06297}, 2017.

\bibitem{monteiro2018conditional}
M.~Monteiro, M.~A. Figueiredo, and A.~L. Oliveira, ``Conditional random fields
  as recurrent neural networks for 3d medical imaging segmentation,''
  \emph{arXiv:1807.07464}, 2018.

\bibitem{grady2006random}
L.~Grady, ``Random walks for image segmentation,'' \emph{IEEE PAMI}, vol.~28,
  no.~11, pp. 1768--1783, 2006.

\bibitem{shaban2017one}
A.~Shaban, S.~Bansal, Z.~Liu, I.~Essa, and B.~Boots, ``One-shot learning for
  semantic segmentation,'' 2017.

\bibitem{zhao2019data}
A.~Zhao, G.~Balakrishnan, F.~Durand, J.~V. Guttag, and A.~V. Dalca, ``Data
  augmentation using learned transformations for one-shot medical image
  segmentation,'' in \emph{CVPR}, 2019, pp. 8543--8553.

\bibitem{chaitanya2019semi}
K.~Chaitanya, N.~Karani, C.~F. Baumgartner, A.~Becker, O.~Donati, and
  E.~Konukoglu, ``Semi-supervised and task-driven data augmentation,'' in
  \emph{IPMI}.\hskip 1em plus 0.5em minus 0.4em\relax Springer, 2019, pp.
  29--41.

\bibitem{liu2018learning}
Y.~Liu, J.~Lee, M.~Park, S.~Kim, E.~Yang, S.~J. Hwang, and Y.~Yang, ``Learning
  to propagate labels: Transductive propagation network for few-shot
  learning,'' \emph{ICLR}, 2019.

\bibitem{feyjie2020semi}
A.~R. Feyjie, R.~Azad, M.~Pedersoli, C.~Kauffman, I.~B. Ayed, and J.~Dolz,
  ``Semi-supervised few-shot learning for medical image segmentation,''
  \emph{arXiv preprint arXiv:2003.08462}, 2020.

\bibitem{asano2019critical}
Y.~M. Asano, C.~Rupprecht, and A.~Vedaldi, ``A critical analysis of
  self-supervision, or what we can learn from a single image,'' \emph{ICLR},
  2020.

\bibitem{guo2019reliable}
L.-Z. Guo, Y.-F. Li, M.~Li, J.-F. Yi, B.-W. Zhou, and Z.-H. Zhou, ``Reliable
  weakly supervised learning: Maximize gain and maintain safeness,''
  \emph{arXiv preprint arXiv:1904.09743}, 2019.

\bibitem{orting2019crowdsourcing}
S.~{\O}rting, A.~Doyle, M.~H.~A. van Hilten, O.~Inel, C.~R. Madan, P.~Mavridis,
  H.~Spiers, and V.~Cheplygina, ``{A survey of crowdsourcing in medical image
  analysis},'' \emph{arXiv:1902.09159}, 2019.

\bibitem{szegedy2015going}
C.~Szegedy, W.~Liu, Y.~Jia, P.~Sermanet, S.~Reed, D.~Anguelov, D.~Erhan,
  V.~Vanhoucke, and A.~Rabinovich, ``{Going deeper with convolutions},'' in
  \emph{CVPR}, 2015, pp. 1--9.

\bibitem{lee2015deeply}
C.-Y. Lee, S.~Xie, P.~Gallagher, Z.~Zhang, and Z.~Tu, ``{Deeply-supervised
  nets},'' in \emph{Artificial Intellig and Stat}, 2015, pp. 562--570.

\bibitem{krebs2019learning}
J.~Krebs, H.~Delingette, B.~Mailh{\'e}, N.~Ayache, and T.~Mansi, ``Learning a
  probabilistic model for diffeomorphic registration,'' \emph{IEEE TMI},
  vol.~38, no.~9, pp. 2165--2176, 2019.

\bibitem{azadi2019semantic}
S.~Azadi, M.~Tschannen, E.~Tzeng, S.~Gelly, T.~Darrell, and M.~Lucic,
  ``{Semantic Bottleneck Scene Generation},'' \emph{arXiv:1911.11357}, 2019.

\bibitem{kurzman2019class}
L.~Kurzman, D.~Vazquez, and I.~Laradji, ``Class-based styling: Real-time
  localized style transfer with semantic segmentation,'' in \emph{ICCV
  Workshops}, 2019, pp. 0--0.

\bibitem{crum2006generalized}
W.~R. Crum, O.~Camara, and D.~L. Hill, ``Generalized overlap measures for
  evaluation and validation in medical image analysis,'' \emph{IEEE TMI},
  vol.~25, no.~11, pp. 1451--1461, 2006.

\end{thebibliography}


\clearpage

\setcounter{section}{0}
\setcounter{table}{0}
\setcounter{figure}{0}

\title{Supplementary Material}
\author{Gabriele~Valvano, Andrea~Leo, Sotirios~A.~Tsaftaris}
\date{}
\maketitle

\section{Dice score and Hausdorff Distance for Single Anatomical Regions}\label{app:organ_dice_haussdorff}
We report Dice score and Hausdorff Distance (HD, in pixels) for each organ of the medical datasets in Table~\ref{tab:multi_organ_metrics_acdc},~\ref{tab:multi_organ_metrics_lvsc},~\ref{tab:multi_organ_metrics_chaost1},~\ref{tab:multi_organ_metrics_chaost2}. Results consider training the segmentors with half of the weakly annotated train set (see Section IV-A). Notice that the average of the Dice score obtained for a method across classes is different from the multi-class Dice score~\cite{crum2006generalized} reported in Table I. In fact, given a multi-class segmentation mask $\mathimage{y}$ and the prediction $\tilde{\mathimage{y}}$:
    $\frac{1}{c} \sum_{i=1}^c \frac{2|\tilde{\mathimage{y}}_i \cdot \mathimage{y}_i|}{|\tilde{\mathimage{y}}_i|+|\mathimage{y}_i|}
    \neq  
    \frac{2|\tilde{\mathimage{y}} \cdot 
    \mathimage{y}|}{|\tilde{\mathimage{y}}|+|\mathimage{y}|}$,
where $i$ refers to each class and $\mathscal{c}$ is the number of classes.

Finally, in Table~\ref{tab:acdc_test_server} we report metrics obtained after training on all the available ACDC data and testing on 50 extra patients using the online evaluation plateform at \url{https://acdc.creatis.insa-lyon.fr/#challenges}.

\begin{table}[h!]
    \caption{Dice score and Haussdorff distance (HD) for single organs in ACDC. Abbreviations are as follows: RV: right ventricle, MYO: myocardium, LV: left ventricle.}
    \label{tab:multi_organ_metrics_acdc}
    \centering
    \setlength{\tabcolsep}{1.2pt}
    \resizebox{\columnwidth}{!}{%
    
    \begin{tabular}{l c|c|c ? c|c|c}
    		& \multicolumn{3}{c}{Dice} & \multicolumn{3}{c}{HD} \\ 
    		\cmidrule(lr){2-4} \cmidrule(lr){5-7}
    		\multicolumn{1}{l}{Model} & RV & MYO & LV & RV & MYO & LV \\
    		\midrule

    		\multicolumn{1}{l}{UNet\textsubscript{PCE}}   
    		                 & 69.3\textsubscript{11}   
    		                 & 76.4\textsubscript{06}    
    		                 & 84.2\textsubscript{07}   
    		                 & ~84.7\textsubscript{29}   
    		                 & ~79.5\textsubscript{23}    
    		                 & ~74.4\textsubscript{28} \\  
    		\multicolumn{1}{l}{UNet\textsubscript{WPCE}}   
    		                 & 56.3\textsubscript{13}   
    		                 & 67.5\textsubscript{06}    
    		                 & 78.4\textsubscript{09}    
    		                 & 120.5\textsubscript{16}   
    		                 & ~99.6\textsubscript{13}    
    		                 & ~97.4\textsubscript{14} \\  
    		\multicolumn{1}{l}{UNet\textsubscript{CRF}}
    		                 & 59.0\textsubscript{14}
    		                 & 66.1\textsubscript{06}
    		                 & 76.6\textsubscript{09}
    		                 & 117.8\textsubscript{20}
    		                 & 103.2\textsubscript{11}
    		                 & ~99.6\textsubscript{13} \\
    		\multicolumn{1}{l}{TS-UNet\textsubscript{CRF}}
    		                 & 27.2\textsubscript{10}
    		                 & 40.8\textsubscript{08}
    		                 & 47.9\textsubscript{12}
    		                 & 133.9\textsubscript{12}
    		                 & 111.9\textsubscript{09}
    		                 & 115.6\textsubscript{10} \\
    		\multicolumn{1}{l}{PostDAE}	
    		                 & 55.6\textsubscript{12}
    		                 & 66.7\textsubscript{07}
    		                 & 80.6\textsubscript{07}
    		                 & 103.4\textsubscript{18}
    		                 & ~88.7\textsubscript{12}
    		                 & ~80.6\textsubscript{15} \\
    		\multicolumn{1}{l}{UNet\textsubscript{D}}
    		                 & 40.4\textsubscript{15}   
    		                 & 59.7\textsubscript{08}    
    		                 & 75.3\textsubscript{09} 
    		                 & ~33.5\textsubscript{10}   
    		                 & ~25.7\textsubscript{12}    
    		                 & ~25.2\textsubscript{14} \\  
    		\multicolumn{1}{l}{ACCL}
    		                 & 73.5\textsubscript{10}   
    		                 & 79.7\textsubscript{05}    
    		                 & 87.8\textsubscript{06} 
    		                 & ~26.1\textsubscript{24}   
    		                 & ~28.8\textsubscript{25}    
    		                 & ~16.6\textsubscript{20} \\  
    		\multicolumn{1}{l}{\hcell{Ours}}   
    		                 & \hcell{75.2\textsubscript{12}}   
    		                 & \hcell{81.7\textsubscript{05}}    
    		                 & \hcell{87.9\textsubscript{05}} 
    		                 & \hcell{~22.7\textsubscript{27}}   
    		                 & \hcell{~26.8\textsubscript{30}}    
    		                 & \hcell{~25.2\textsubscript{27}} \\  
    		\bottomrule\\
    	\end{tabular}
    }
\end{table}

\begin{table}[h!]
    \caption{Dice score and Haussdorff distance (HD) for single organs in LVSC. MYO stands for myocardium.}
    \label{tab:multi_organ_metrics_lvsc}
    \centering
    \setlength{\tabcolsep}{1.2pt}
    
    \begin{tabular}{l c ? c}
    		& \multicolumn{1}{c}{Dice} & \multicolumn{1}{c}{HD} \\ 
    		\cmidrule(lr){2-2} \cmidrule(lr){3-3}
    		\multicolumn{1}{l}{Model} & MYO & MYO \\
    		\midrule

    		\multicolumn{1}{l}{UNet\textsubscript{PCE}}   
    		                 & 62.3\textsubscript{09}  
    		                 & 55.7\textsubscript{28}  \\  
    		\multicolumn{1}{l}{UNet\textsubscript{WPCE}}   
    		                 & 59.1\textsubscript{07}  
    		                 & 52.4\textsubscript{23} \\  
    		\multicolumn{1}{l}{UNet\textsubscript{CRF}}  
    		                 & 60.4\textsubscript{08}
    		                 & 53.0\textsubscript{27} \\  
    		\multicolumn{1}{l}{TS-UNet\textsubscript{CRF}}  
    		                 & 50.5\textsubscript{07}
    		                 & 93.4\textsubscript{27}\\  
    		\multicolumn{1}{l}{PostDAE}
    		                 & 58.6\textsubscript{07} 
    		                 & 47.5\textsubscript{22} \\  
    		\multicolumn{1}{l}{UNet\textsubscript{D}}
    		                 & 31.7\textsubscript{09} 
    		                 & 44.7\textsubscript{20} \\  
    		\multicolumn{1}{l}{ACCL}
    		                 & 65.9\textsubscript{08} 
    		                 & 24.0\textsubscript{19} \\  
    		\multicolumn{1}{l}{\hcell{Ours}}   
    		                 & \hcell{65.5\textsubscript{08}}
    		                 & \hcell{27.5\textsubscript{25}}  \\  
    		\bottomrule\\
    	\end{tabular}
\end{table}

\begin{table}[h!]
    \caption{Dice score and Haussdorff distance (HD) for single organs in CHAOS-T1. Abbreviations are as follows: L: liver, RK: right kidney, LK: left kidney, S: spleen.}
    \label{tab:multi_organ_metrics_chaost1}
    \centering
    \setlength{\tabcolsep}{1.2pt}
    
    \resizebox{\columnwidth}{!}{%
    \begin{tabular}{l c|c|c|c ? c|c|c|c}
    		& \multicolumn{4}{c}{Dice} & \multicolumn{4}{c}{HD} \\ 
    		\cmidrule(lr){2-5} \cmidrule(lr){6-9}
    		\multicolumn{1}{l}{Model} & L & RK & LK & S & L & RK & LK & S  \\
    		\midrule

    		\multicolumn{1}{l}{UNet\textsubscript{PCE}}   
    		                 & 43.5\textsubscript{07}
    		                 & 21.3\textsubscript{04}
    		                 & ~9.1\textsubscript{03}    
    		                 & 25.9\textsubscript{07}
    		                 & 133.5\textsubscript{01}    
    		                 & 157.1\textsubscript{04} 
    		                 & 151.7\textsubscript{01}    
    		                 & 133.8\textsubscript{07} \\  
    		\multicolumn{1}{l}{UNet\textsubscript{WPCE}}   
    		                 & 42.5\textsubscript{09} 
    		                 & 29.2\textsubscript{02}
    		                 & 16.6\textsubscript{02}    
    		                 & 25.7\textsubscript{05}
    		                 & 121.3\textsubscript{01}    
    		                 & 114.5\textsubscript{03} 
    		                 & 154.6\textsubscript{01}    
    		                 & 128.7\textsubscript{01} \\  
    		\multicolumn{1}{l}{UNet\textsubscript{CRF}}
    		                 & 37.3\textsubscript{09}
    		                 & 20.0\textsubscript{06}
    		                 & 16.3\textsubscript{04}
    		                 & 27.9\textsubscript{13}
    		                 & 119.2\textsubscript{10}
    		                 & 148.9\textsubscript{04}
    		                 & 148.4\textsubscript{06}
    		                 & 101.8\textsubscript{08} \\
    		\multicolumn{1}{l}{TS-UNet\textsubscript{CRF}}
    		                 & 41.1\textsubscript{12}
    		                 & 13.2\textsubscript{04}
    		                 & ~6.2\textsubscript{02}
    		                 & 16.5\textsubscript{06}
    		                 & 110.6\textsubscript{18}
    		                 & 157.7\textsubscript{04}
    		                 & 153.8\textsubscript{05}
    		                 & 163.8\textsubscript{08} \\
    		\multicolumn{1}{l}{PostDAE}
    		                 & 32.8\textsubscript{07} 
    		                 & 57.9\textsubscript{07} 
    		                 & 57.1\textsubscript{06} 
    		                 & 58.4\textsubscript{11}
    		                 & 100.1\textsubscript{13}    
    		                 & 192.0\textsubscript{00}    
    		                 & 184.8\textsubscript{06}    
    		                 & 192.0\textsubscript{00} \\   
    		\multicolumn{1}{l}{UNet\textsubscript{D}}
    		                 & 60.2\textsubscript{05} 
    		                 & 46.4\textsubscript{10}
    		                 & 46.9\textsubscript{06}    
    		                 & 41.3\textsubscript{12}
    		                 & ~59.9\textsubscript{02}    
    		                 & ~93.5\textsubscript{37}    
    		                 & 151.0\textsubscript{03}    
    		                 & 123.1\textsubscript{03} \\   
    		\multicolumn{1}{l}{ACCL}
    		                 & 65.0\textsubscript{12}
    		                 & 57.3\textsubscript{06}
    		                 & 49.4\textsubscript{09}    
    		                 & 51.2\textsubscript{14}
    		                 & ~35.3,\textsubscript{12}    
    		                 & 178.3\textsubscript{19}    
    		                 & ~85.0,\textsubscript{04}    
    		                 & 100.9\textsubscript{18} \\   
    		\multicolumn{1}{l}{\hcell{Ours}}   
    		                 & \hcell{64.0\textsubscript{07}} 
    		                 & \hcell{68.5\textsubscript{06}}
    		                 & \hcell{59.6\textsubscript{09}}    
    		                 & \hcell{39.7\textsubscript{08}}
    		                 & \hcell{~62.0\textsubscript{05}}    
    		                 & \hcell{~27.4\textsubscript{13}}   
    		                 & \hcell{~34.7\textsubscript{03}}    
    		                 & \hcell{~60.8\textsubscript{27}} \\  
    		\bottomrule\\
    	\end{tabular}
    }
\end{table}

\begin{table}[h!]
    \caption{Dice score and Haussdorff distance (HD) for single organs in CHAOS-T2. Abbreviations are as follows: L: liver, RK: right kidney, LK: left kidney, S: spleen.}
    \label{tab:multi_organ_metrics_chaost2}
    \centering
    \setlength{\tabcolsep}{1.2pt}
    
    \resizebox{\columnwidth}{!}{%
    \begin{tabular}{l c|c|c|c ? c|c|c|c}
    		& \multicolumn{4}{c}{Dice} & \multicolumn{4}{c}{HD} \\ 
    		\cmidrule(lr){2-5} \cmidrule(lr){6-9}
    		\multicolumn{1}{l}{Model} & L & RK & LK & S & L & RK & LK & S  \\
    		\midrule

    		\multicolumn{1}{l}{UNet\textsubscript{PCE}}   
    		                 & 48.4\textsubscript{08}
    		                 & 23.9\textsubscript{05}
    		                 & ~9.7\textsubscript{02}    
    		                 & 27.7\textsubscript{07}
    		                 & 133.1\textsubscript{01}    
    		                 & 155.9\textsubscript{04} 
    		                 & 151.3\textsubscript{01}    
    		                 & 114.6\textsubscript{09} \\ 
    		\multicolumn{1}{l}{UNet\textsubscript{WPCE}}   
    		                 & 55.6\textsubscript{09} 
    		                 & 31.5\textsubscript{04}
    		                 & 28.4\textsubscript{03}    
    		                 & 32.2\textsubscript{10}
    		                 & 106.0\textsubscript{09}    
    		                 & 129.9\textsubscript{04} 
    		                 & 135.9\textsubscript{02}    
    		                 & 101.0\textsubscript{03} \\  
    		\multicolumn{1}{l}{UNet\textsubscript{CRF}}
    		                 & 48.0\textsubscript{09} 
    		                 & 26.4\textsubscript{15} 
    		                 & 19.9\textsubscript{03} 
    		                 & 32.9\textsubscript{12}
    		                 & 117.2\textsubscript{11}    
    		                 & 151.1\textsubscript{04}    
    		                 & 141.1\textsubscript{09}    
    		                 & ~91.0\textsubscript{11} \\   
    		\multicolumn{1}{l}{TS-UNet\textsubscript{CRF}}
    		                 & 44.5\textsubscript{10} 
    		                 & ~7.0\textsubscript{03} 
    		                 & ~6.4\textsubscript{03} 
    		                 & 18.4\textsubscript{05}
    		                 & ~90.8\textsubscript{14}    
    		                 & 157.6\textsubscript{04}    
    		                 & 154.5\textsubscript{05}    
    		                 & 157.3\textsubscript{08} \\   
    		\multicolumn{1}{l}{PostDAE}
    		                 & 43.4\textsubscript{07} 
    		                 & 57.9\textsubscript{07} 
    		                 & 57.5\textsubscript{06} 
    		                 & 58.4\textsubscript{11}
    		                 & ~76.4\textsubscript{12}    
    		                 & 192.0\textsubscript{00}    
    		                 & 190.2\textsubscript{03}    
    		                 & 192.0\textsubscript{00} \\
    		\multicolumn{1}{l}{UNet\textsubscript{D}}
    		                 & 63.6\textsubscript{04} 
    		                 & 53.0\textsubscript{10}
    		                 & 45.0\textsubscript{08}    
    		                 & 34.1\textsubscript{10}
    		                 & ~52.8\textsubscript{06}    
    		                 & 127.7\textsubscript{23}    
    		                 & 108.5\textsubscript{01}    
    		                 & 113.0\textsubscript{06} \\   
    		\multicolumn{1}{l}{ACCL}
    		                 & 63.2\textsubscript{10} 
    		                 & 42.8\textsubscript{10}
    		                 & 46.5\textsubscript{09}    
    		                 & 56.5\textsubscript{12}
    		                 & ~47.3\textsubscript{18}    
    		                 & ~77.4\textsubscript{36}    
    		                 & ~94.7\textsubscript{29}    
    		                 & ~98.3\textsubscript{44} \\   
    		\multicolumn{1}{l}{\hcell{Ours}}   
    		                 & \hcell{56.3\textsubscript{06}}
    		                 & \hcell{68.6\textsubscript{07}}
    		                 & \hcell{61.4\textsubscript{09}}   
    		                 & \hcell{44.2\textsubscript{08}}
    		                 & \hcell{~65.7\textsubscript{03}}    
    		                 & \hcell{~44.6\textsubscript{12}}   
    		                 & \hcell{~40.0\textsubscript{18}}    
    		                 & \hcell{~63.6\textsubscript{27}} \\  
    		\bottomrule\\
    	\end{tabular}
    }

\end{table}

\begin{table}[h!]
    \caption{Dice score and Haussdorff distance (HD) of the proposed approach trained on all the available ACDC data, and tested on 50 extra patients using the challenge server. Note that the server does not provide information about the standard deviation, nor a higher precision for the Dice score. Abbreviations are as follows: RV: right ventricle, MYO: myocardium, LV: left ventricle.}
    \label{tab:acdc_test_server}
    \centering
    \setlength{\tabcolsep}{1.2pt}
    
    \begin{tabular}{l c|c|c ? c|c|c}
    		& \multicolumn{3}{c}{Dice} & \multicolumn{3}{c}{HD} \\ 
    		\cmidrule(lr){2-4} \cmidrule(lr){5-7}
    		\multicolumn{1}{l}{Cardiac Phase~~} & ~RV~ & MYO & ~LV~ & ~RV~ & MYO & ~LV~ \\
    		\midrule

    		\multicolumn{1}{l}{End-diastole}   
    		                 & 89
    		                 & 81
    		                 & 93
    		                 & 16.6
    		                 & 45.3
    		                 & 20.9 \\  
    		\multicolumn{1}{l}{End-systole}
    		                 & 84
    		                 & 84
    		                 & 88
    		                 & 20.3
    		                 & 44.2 
    		                 & 27.1 \\
    		\bottomrule\\
    	\end{tabular}
\end{table}

\FloatBarrier
\section{Fully Supervised Learning}\label{app:fully_sup}
We conducted experiments to analyse model performance when it is trained with mask supervision rather than with scribbles. We report results in Table~\ref{tab:full_supervision}. As can be seen from the table, the same model works well with full supervision, and it improves performance when training with masks, rather than when using only scribble annotations. 

We highlight that we conducted these experiments while keeping exactly the \textit{same framework and hyperparameters}. It is possible that the choice of better hyperparameters could further improve the reported numbers (for example, changing the learning rate). However, since our scope is not related to training with full supervision, we don’t investigate this further.

\begin{table}[h!]
    \caption{Training our method with scribbles and with mask supervision. We report the Dice average (standard deviation as subscript) obtained on the test data for each dataset.}
    \label{tab:full_supervision}
    \centering
    \setlength{\tabcolsep}{1.5pt}
    
    \resizebox{\columnwidth}{!}{%
    \begin{tabular}{l c|c|c|c|c}
    		\multicolumn{1}{l}{}
    		& \multicolumn{5}{c}{Dataset}
    		\\ \cmidrule(lr){2-6}
    		
    		Supervision ~~ & ACDC & LVSC & CHAOS-T1 & CHAOS-T2 & PPSS \\
    		\cmidrule[\arrayrulewidth]{1-6}
    		Scribbles 
    		                 & 84.3\textsubscript{04}
    		                 & 65.5\textsubscript{08}
    		                 & 56.8\textsubscript{05} 
    		                 & 57.8\textsubscript{04}
    		                 & 74.6\textsubscript{04} \\
    		Masks
    		                 & 84.3\textsubscript{02}
    		                 & 68.8\textsubscript{07}
    		                 & 65.7\textsubscript{03} 
    		                 & 65.9\textsubscript{02}
    		                 & 76.9\textsubscript{04} \\
    		\bottomrule
    	\end{tabular}
    }
\end{table}

\section{Additional Figures}\label{app:figures}
\begin{figure}[h!]
    \centering
    \includegraphics[width=0.48\textwidth]{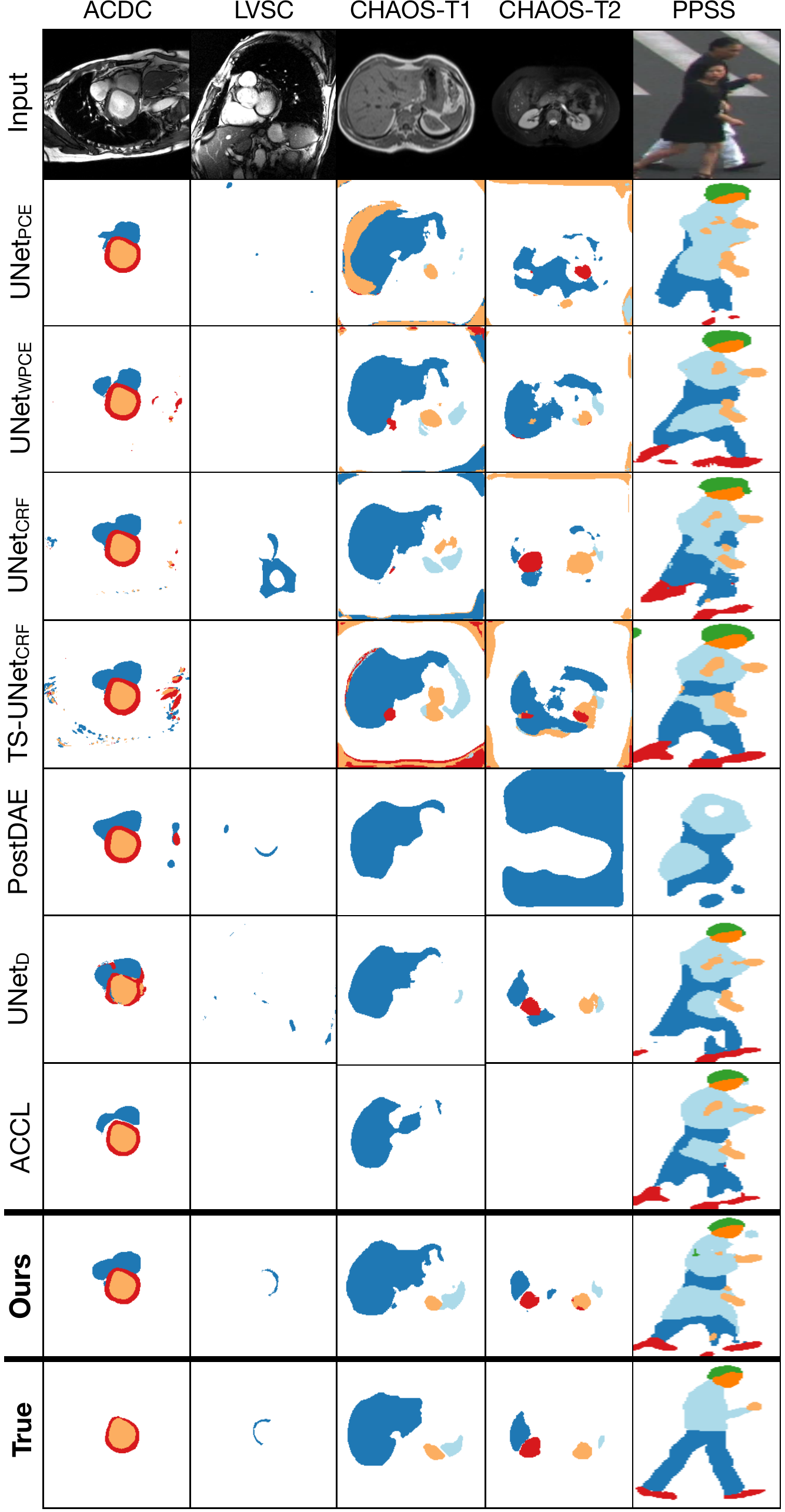}
    \caption{Example of model failures. In both ACDC and LVSC, the apical and the basal slices of the heart are the hardest to segment, due to intrinsic uncertainty of the cardiac boundaries, resulting in over/under-segmentations in all the models.
    For CHAOS, we show that all models make mistakes when the organ boundaries have low contrast, though our model preserves realistic outputs. In PPSS, we show that occlusions make the segmentation task harder; for example, if two people overlap, all model will try to segment both people, rather than only one.}
    \label{fig:example_failures}
\end{figure}

We report examples of segmentation failures for the proposed approach and for the benchmark models in Fig.~\ref{fig:example_failures}.

\end{document}